\documentclass[10pt,twocolumn,letterpaper]{article}
\usepackage[T1]{fontenc}
\usepackage[pagenumbers]{cvpr}
\usepackage{multirow}
\usepackage{array}
\usepackage{placeins}
\usepackage{tabularx}
\newcolumntype{P}[1]{>{\raggedright\arraybackslash}p{#1}}\newcolumntype{Y}{>{\raggedright\arraybackslash}X}

\newcommand{\labels}{\mathcal{Y}}
\newcommand{\chosen}{y_+}
\newcommand{\rejected}{y_-}

\newcommand{\pmval}[2]{\begin{tabular}[c]{@{}c@{}}$#1$\\[-2pt]\scriptsize$\pm#2$\end{tabular}}

\definecolor{cvprblue}{rgb}{0.21,0.49,0.74}
\usepackage[pagebackref,breaklinks,colorlinks,allcolors=cvprblue]{hyperref}

\def\confName{CVPR}
\def\confYear{2027}

\title{Lightweight Adaptation of General-Purpose VLMs\\for Multispectral and SAR Image Understanding}
\author{Shanji Liu\quad kelu Yao\quad Junxiao Xue\quad Chenchen Sun\quad Yekai Huang\\Hongkun Pan\quad Chenghui Lv\quad Xiangyang Miao\quad Yaying Chen\quad Chao Li}

\begin{document}
\maketitle

\begin{abstract}
Adapting general-purpose vision-language models (VLMs) to multispectral imagery (MSI) and synthetic aperture radar (SAR) requires more than compatible inputs: rendered sensor patterns differ from natural images, and predicted classes can contradict stated evidence. We propose Multiview Sensor Adaptation (MVSA), a lightweight framework that retains the pretrained architecture and its native multi-image interface. MVSA renders aligned observations into six three-channel views, with physical names identifying their spectral or radar content in the prompt. Joint visual and language low-rank adaptation (LoRA) trains the model to predict classes and supporting evidence from these views. To target class omissions despite supporting evidence, evidence-consistency direct preference optimization (EC-DPO) contrasts complete responses with alternatives that omit one true class while retaining identical evidence. On BigEarthNet-v2, adaptation improves Qwen3-VL micro F1 from 0.5921 to 0.8526 under the same six-view interface; EC-DPO increases exact match from 0.4288 to 0.4720 over joint supervised fine-tuning. Controlled experiments show that shared bands can mask reliance on derived views, while sensor interventions reveal greater SAR reliance under clouds. Experiments across six VLM backbones, flood verification, and dense captioning evaluate the interface's broader applicability.

\end{abstract}

\begin{figure}[t]
\centering
\includegraphics[width=\linewidth]{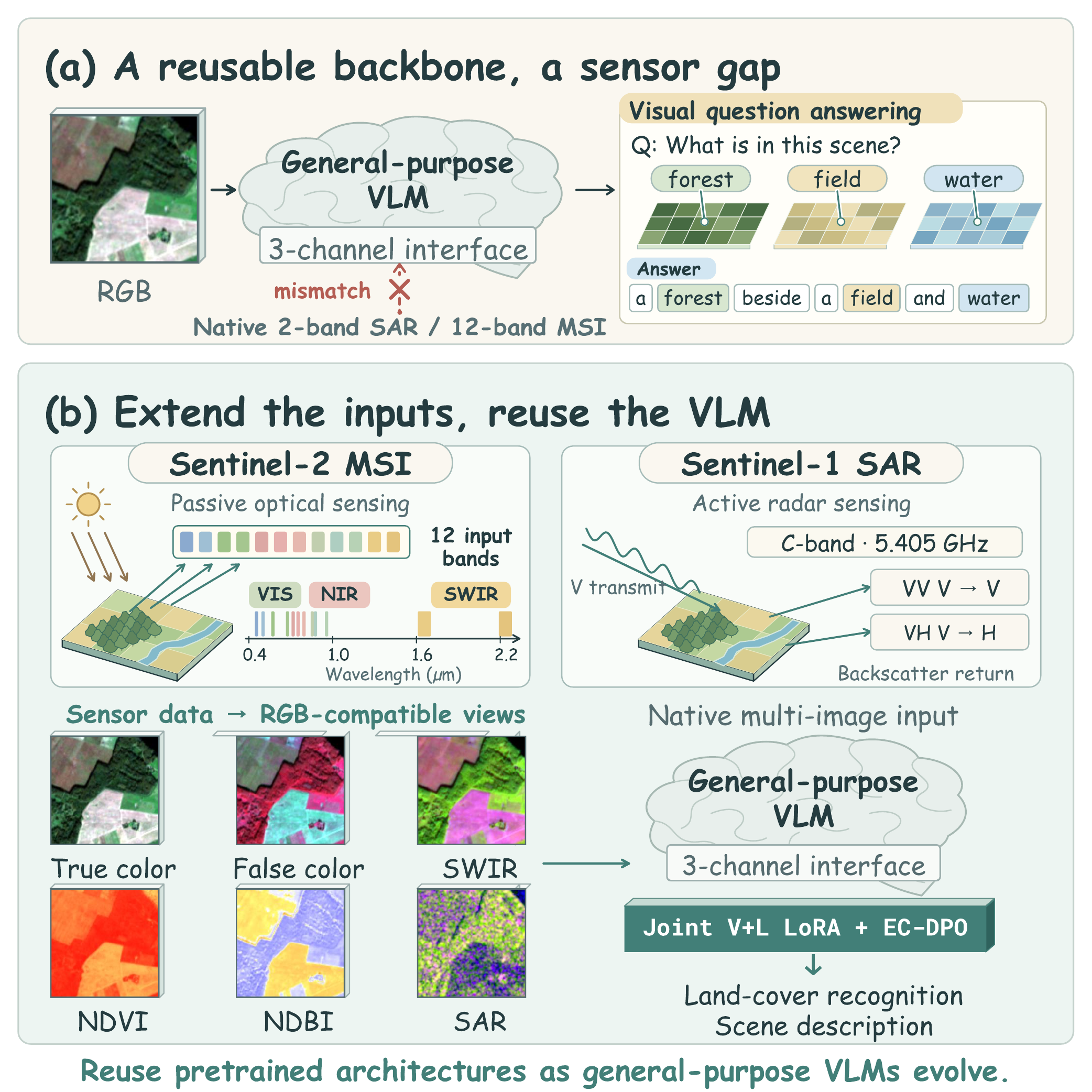}
\caption{Adapting general-purpose VLMs to MSI and SAR observations. (a) General-purpose VLMs expect natural RGB inputs, leaving a sensor gap for native MSI and SAR. (b) Rendering aligned bands into six named views enables lightweight adaptation via the native image interface. Sensor diagrams and the visual question answering (VQA) example are schematic; all six views share one patch.}
\label{fig:teaser}
\end{figure}

\section{Introduction}

Joint optical and radar observations support land-cover interpretation through complementary measurements. Multispectral imagery (MSI) captures visible and infrared reflectance, while synthetic aperture radar (SAR) provides backscatter observations even when clouds obscure optical imagery \citep{bigearthnetmm2021,reben2024}. General-purpose vision-language models (VLMs) combine recognition with language generation \citep{flamingo2022,qwen25vl2025,qwen3vl2025}, offering reusable backbones for interpreting these observations and describing supporting sensor cues. However, their RGB encoders cannot directly ingest raw MSI and SAR tensors.

Specialist encoders process native MSI and SAR tensors but require task-specific heads for downstream prediction \citep{satmae2022,croma2023,terrafm2025}. Dedicated remote-sensing VLMs use custom input modules or domain pretraining \citep{geochat2024,earthdial2024,earthonevision2026}, which can complicate adoption of newer backbones. Rendering band combinations as three-channel images preserves the existing interface \citep{zeroshot2025multispectral,unlocking2026multispectral}, but input compatibility alone does not ensure effective sensor interpretation. We ask: \emph{how can general-purpose VLMs jointly interpret MSI and SAR with lightweight adaptation, retaining their native image interface while producing consistent class and evidence outputs?}

We propose Multiview Sensor Adaptation (MVSA), combining a named multisensor interface with joint visual--language adaptation (\cref{fig:teaser}). To make inputs compatible, MVSA renders each aligned observation into six views: true color, false color, short-wave infrared (SWIR), normalized difference vegetation index (NDVI), normalized difference built-up index (NDBI), and SAR backscatter. Yet the same rendered color can represent different physical quantities across views. We therefore attach physical names in a fixed prompt order, identifying each view's spectral or radar content through the VLM's native multi-image interface.

Compatible inputs still differ from the natural images used in pretraining: spectral composites, index maps, and radar speckle have distinct visual statistics. Language-only tuning leaves the visual representation fixed. We train low-rank adapters (LoRA) \citep{lora2022} in both visual blocks and language layers, keeping pretrained weights frozen. Joint supervised fine-tuning (SFT) learns land-cover classes followed by concise supporting evidence. Parameter-matched, three-seed comparisons test the benefit of visual adaptation beyond increasing language-adapter capacity.

An SFT response may omit forest from its class list while describing forest canopy. Standard SFT does not explicitly contrast complete responses with such inconsistent alternatives. We introduce evidence-consistency direct preference optimization (EC-DPO), pairing a complete reference response with an alternative that removes exactly one true class while retaining identical evidence, including a cue specific to that class. This controlled contrast targets omissions under fixed evidence and trains the policy to favor consistent class--evidence outputs.

Shared spectral content can compensate for an omitted view, so aggregate accuracy alone cannot reveal view reliance or class-specific errors. On BigEarthNet-v2 \citep{reben2024}, adaptation raises Qwen3-VL micro F1 from 0.5921 to 0.8526 using the same six-view interface. Relative to joint SFT, EC-DPO improves exact match from 0.4288 to 0.4720 and wetland recall from 0.7481 to 0.8010. View ablations and donor swaps reveal how shared bands mask derived-view reliance in aggregate metrics while minority-class recall remains sensitive. Sensor interventions test image dependence and show increased SAR reliance under clouds; blinded human audits assess evidence plausibility. Evaluations across six VLM backbones, Sen1Floods11 flood verification \citep{sen1floods11}, and BigEarthNet.txt dense captioning \citep{bigearthnettxt2026} assess broader applicability.

Our main contributions are threefold:
\begin{itemize}
\item \textbf{(i) Multiview Sensor Adaptation (MVSA):} We adapt frozen VLMs to aligned MSI and SAR inputs, where we combine six named views with joint LoRA.
\item \textbf{(ii) Evidence-Consistency Preference Optimization:} We analyze label and evidence inconsistency after SFT, and we introduce EC-DPO, which contrasts complete responses against class omissions under shared evidence.
\item \textbf{(iii) Rendering and Modality Analysis:} Through rendering ablations, view omission, donor replacement, cloud occlusion tests, and double-blind expert audits, we show how spectral redundancy, minority class trade-offs, and radar reliance under clouds govern predictions.
\end{itemize}

\section{Related Work}

\paragraph{Remote-sensing foundation models and VLMs.}
Self-supervised remote-sensing encoders learn multispectral, radar, and temporal features via contrastive learning and masked modeling \citep{seco2021,satmae2022,scalemae2023,ssl4eos12_2023,spectralgpt2024,prithvi2024,skysense2024,croma2023,anysat2025,terrafm2025}. Although these native-tensor specialists yield strong discriminative features, they require task-specific heads and lack open-ended language interaction. Conversely, building on general-purpose VLMs \citep{instructblip2023,llava2023,qwen25vl2025,qwen3vl2025} and vision-language retrieval models \citep{remoteclip2024,rs5m2024}, dedicated remote-sensing VLMs such as GeoChat \citep{geochat2024}, EarthDial \citep{earthdial2024}, and EarthMind \citep{earthmind2025} attach custom projectors, or they pretrain on satellite corpora. However, domain pretraining demands heavy compute, struggles with multi-label formatting, and lags behind newer general-purpose backbones. While native-tensor encoders serve as discriminative upper references, we adapt frozen general-purpose VLMs without architectural changes.

\paragraph{Multi-view multispectral interfaces.}
Llama3-MS-CLIP adapts CLIP's patch embeddings through multispectral contrastive pretraining \citep{marimo2025beyond}. Rendering-based approaches instead preserve the RGB interface, mapping Sentinel-2 bands to true-color, false-color, and index views for training-free inference \citep{zeroshot2025multispectral,unlocking2026multispectral}. Sentinel2Cap evaluates RGB, multispectral, and SAR pseudo-RGB inputs with modality-specific prompts for zero-shot captioning \citep{tosato2026sentinel2cap}. These studies motivate sensor-aware interfaces but do not isolate how joint visual--language adaptation interacts with redundant bands. We combine six physically named MSI--SAR views with parameter-matched tuning and counterfactual interventions, examining how visual adaptation and shared bands govern minority-class errors and resilience under cloud cover.

\begin{figure*}[!t]
\centering
\includegraphics[width=\textwidth]{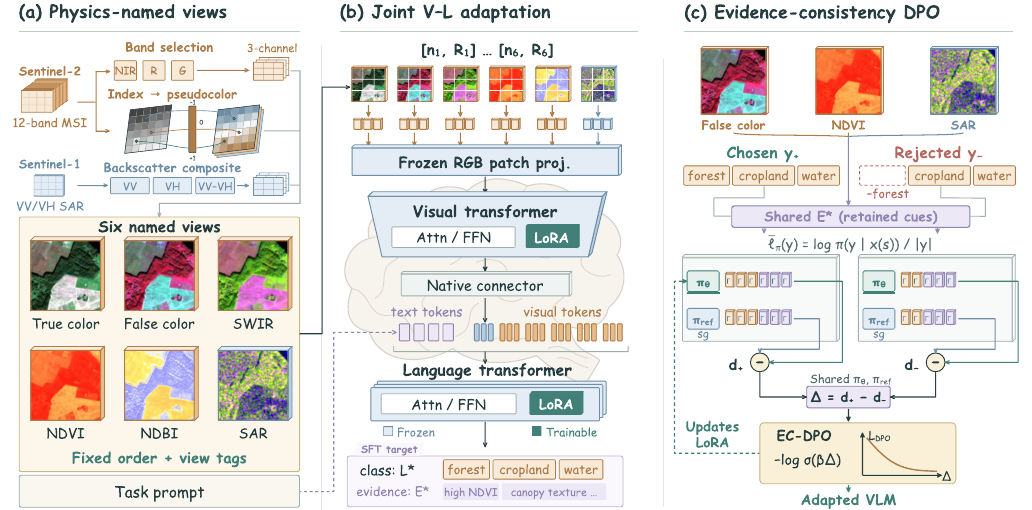}
\caption{Overview of MVSA. (a) Aligned MSI and SAR observations are rendered into six physically named views. (b) Joint visual and language LoRA learns to generate classes and supporting evidence while pretrained weights remain frozen. (c) EC-DPO contrasts complete responses with rejected responses that omit one true class while retaining identical evidence. The policy is initialized from SFT and trained against a frozen reference.}
\label{fig:method}
\end{figure*}

\paragraph{Preference optimization and evidence consistency.}
Parameter-efficient tuning via LoRA, domain post-training, and direct preference optimization (DPO) align language and multimodal models \citep{lora2022,adaptllm2025,dpo2023}. Multimodal DPO variants further reduce visual neglect, where preferences are conditioned on image perturbations \citep{mdpo2024}. Standard preference objectives, however, rank complete responses by overall accuracy or style, so they do not penalize responses that omit a true class while retaining its optical and radar cues. We introduce evidence-consistency DPO (EC-DPO), which constructs rejected responses by dropping one true class while keeping the evidence fixed, coupling class recovery with grounded evidence generation.

\section{Method}
\label{sec:method}

\Cref{fig:method} presents MVSA, which combines six physically named views, joint visual and language adaptation, and subsequent EC-DPO optimization.

\begin{table*}[!t]
\centering
\caption{Main comparison on BigEarthNet-v2 ($3{,}234$ validation patches). $^\dagger$Discriminative encoders take native Sentinel-1/2 (S1+S2) tensors or rendered views without text. $^\S$Lexical check (\texttt{Ev.\ audit}: optical/SAR coverage, attribution, cue support) is N/A (--) for RGB-only models. $^\ddagger$Tier~3 runs a cross-backbone wrapper; Tier~4 and \cref{tab:rendering-design} run the native processor. Bold/underline: best/second-best generative models.}
\label{tab:main}
\footnotesize
\setlength{\tabcolsep}{3.2pt}
\begin{tabular*}{\textwidth}{@{\extracolsep{\fill}}p{0.25\textwidth}p{0.22\textwidth}rrrrrr@{}}
\toprule
Method & Input \& adaptation scope & Trainable & Micro F1 & Macro F1 & Exact match & Format valid & Ev.\ audit \\
\midrule
\multicolumn{8}{l}{\textit{Tier 1: Frozen specialist and visual encoders (non-generative upper references, $n=3{,}234$)$^\dagger$}}\\
TerraFM + linear head \citep{terrafm2025} & native S1+S2 tensors, head only & head & 0.8442 & 0.7725 & 0.4400 & -- & -- \\
CROMA + linear head \citep{croma2023} & native S1+S2 tensors, head only & head & 0.8334 & 0.7509 & 0.4190 & -- & -- \\
ResNet50 + linear head \citep{resnet2016} & six rendered views, head only & head & 0.8012 & 0.7179 & 0.3510 & -- & -- \\
Frozen Qwen3-VL + linear head & six rendered views, head only & head & 0.7904 & 0.7039 & 0.3108 & -- & -- \\
\midrule
\multicolumn{8}{l}{\textit{Tier 2: External remote-sensing VLMs (zero-shot interface diagnostics, $n=3{,}234$)$^\S$}}\\
SkySenseGPT \citep{skysensegpt2024} & true color only, zero-shot & 0 & 0.5925 & 0.4135 & 0.1679 & 0.0\% & -- \\
GeoChat \citep{geochat2024} & true color only, zero-shot & 0 & 0.5530 & 0.3941 & 0.1224 & 13.8\% & -- \\
EarthDial-RGB \citep{earthdial2024} & six rendered views, zero-shot & 0 & 0.2834 & 0.2648 & 0.1259 & 0.0\% & 0.0000 \\
EarthMind \citep{earthmind2025} & six rendered views, zero-shot & 0 & 0.1679 & 0.1485 & 0.0090 & 0.0\% & 0.0000 \\
\midrule
\multicolumn{8}{l}{\textit{Tier 3: General-purpose VLMs, zero-shot and fixed language-LoRA cross-backbone harness ($n=3{,}234$)$^\ddagger$}}\\
Qwen3-VL-8B (zero-shot) \citep{qwen3vl2025} & six named views, zero-shot & 0 & 0.5921 & 0.5107 & 0.1351 & \textbf{100.0\%} & 0.1531 \\
InternVL3.5-8B (zero-shot) \citep{internvl35_2025} & six named views, zero-shot & 0 & 0.5537 & 0.4733 & 0.1184 & \textbf{100.0\%} & 0.1240 \\
Idefics3-8B (zero-shot) \citep{idefics3_2024} & six named views, zero-shot & 0 & 0.5374 & 0.3418 & 0.0965 & \textbf{100.0\%} & 0.0894 \\
SmolVLM2-2.2B (zero-shot) \citep{smolvlm2025} & six named views, zero-shot & 0 & 0.5026 & 0.4539 & 0.0816 & \textbf{100.0\%} & 0.0612 \\
Qwen2.5-VL-7B (SFT) \citep{qwen25vl2025} & six named views, language LoRA & 43.65M & 0.8026 & 0.7200 & 0.3377 & \textbf{100.0\%} & 0.9153 \\
AdaptLLM-RS-3B (SFT) \citep{adaptllm2025} & six named views, language LoRA & 21.82M & 0.7917 & 0.7017 & 0.3380 & \textbf{100.0\%} & 0.9465 \\
InternVL3.5-8B (SFT) \citep{internvl35_2025} & six named views, language LoRA & 43.65M & 0.7907 & 0.6908 & 0.3268 & \textbf{100.0\%} & 0.9422 \\
Idefics3-8B (SFT) \citep{idefics3_2024} & six named views, language LoRA & 41.94M & 0.7903 & 0.7039 & 0.3163 & \textbf{100.0\%} & 0.9397 \\
Qwen3-VL-8B (SFT) \citep{qwen3vl2025} & six named views, language LoRA & 43.65M & 0.7681 & 0.6864 & 0.2919 & \textbf{100.0\%} & 0.9712 \\
SmolVLM2-2.2B (SFT) \citep{smolvlm2025} & six named views, language LoRA & 18.09M & 0.7268 & 0.6197 & 0.2186 & \textbf{100.0\%} & 0.9326 \\
\midrule
\multicolumn{8}{l}{\textit{Tier 4: Ours (Native multi-image grid processor + joint vision-language adaptation, $n=3{,}234$)}}\\
Qwen2.5-VL-7B, Joint SFT & six named views, joint V+L LoRA & 51.52M & 0.8184 & 0.7465 & 0.4063 & \textbf{100.0\%} & 0.9682 \\
Qwen3-VL-8B, Joint SFT & six named views, joint V+L LoRA & 51.35M & \underline{0.8242} & \underline{0.7561} & \underline{0.4288} & \textbf{100.0\%} & \underline{0.9753} \\
\textbf{Qwen3-VL-8B, + EC-DPO (Ours)} & six named views, joint V+L LoRA & 51.35M & \textbf{0.8526} & \textbf{0.7783} & \textbf{0.4720} & \textbf{100.0\%} & \textbf{0.9829} \\
\bottomrule
\end{tabular*}
\end{table*}

\subsection{Rendering Aligned Sensor Observations}
\label{sec:sensor}

Each observation $s=(S^{(2)},S^{(1)})$ pairs Sentinel-2 MSI with Sentinel-1 SAR backscatter. Deterministic operators $\{R_k\}_{k=1}^{6}$ render six three-channel views while preserving the pretrained RGB encoder. These views comprise true color, near-infrared (NIR) false color, SWIR, NDVI, NDBI, and SAR. Across the six views, key optical bands are shared: Red (B04) and NIR (B08) appear in four views, while SWIR (B11) appears in two. Blue (B02) appears only in true color, while Green (B03) is shared by true and false color. The vertical-vertical (VV) and vertical-horizontal (VH) polarizations appear only in SAR. Rendering spectral indices, such as NDVI and NDBI, along with SWIR composites, exposes band ratios before visual encoding, while shared bands provide overlapping cues (Appendix~\ref{app:render}).

The native processor accepts six views within one context window. Since all views pass through one RGB patch encoder, identical pixel colors represent different physical quantities across views. High red intensity, for instance, indicates visible reflectance in true color, dense vegetation in false color, and dry or built-up surfaces in SWIR. Rendered colors alone therefore cannot identify the underlying wavelengths, nor can they distinguish radar polarizations. In~\eqref{eq:rendered-interface}, we attach a physical name $n_k$ to each view $R_k(s)$, following a fixed order within the prompt $p$:
\begin{equation}
\label{eq:rendered-interface}
x(s)=p\mathbin{\Vert}\big\Vert_{k=1}^{6}
\bigl[n_k,\,R_k(s)\bigr],
\end{equation}
where $\Vert$ denotes sequence concatenation. This fixed order tags each sensor source, aiding both classification and image captioning tasks.

\subsection{Class and Evidence Outputs with Joint Adaptation}
\label{sec:instruction}

We merge 18 BigEarthNet-v2 categories, forming six land-cover classes (Supplementary Table~\ref{tab:label-map}): $\labels=\{\text{urban},\ \text{forest},\ \text{cropland},\ \text{grass\_shrub},\ \text{water},\ \text{wetland}\}$. Standard multilabel supervision specifies which classes are present, whereas it does not train the model to explain its predictions. Given label set $L^\star\subseteq\labels$, with audited evidence $E^\star$, outputs serialize as $y=\operatorname{ser}(\hat{L},\hat{E})$ over two lines. The first line lists predicted classes $\hat{L}\subseteq\labels$ after \texttt{class:}. The second provides a concise explanation $\hat{E}$ after \texttt{evidence:}, citing supporting optical and radar cues. For each class $c\in\labels$, we maintain an audited cue lexicon $T_c$, along with its exclusive subset $T_c^\star = T_c \setminus \bigcup_{c'\neq c} T_{c'}$ containing cues that occur in no other class lexicon. These lexicons guide target construction, and they drive automated lexical auditing of sensor coverage, attribution, and cue support, complementing human visual evaluation (Appendix~\ref{app:dict}).

Adaptation must account for rendered view statistics, as well as the structured output format. Pretrained RGB visual blocks are tuned to natural colors and standard image edges. When applied directly to pseudocolor index maps, SWIR moisture composites, and multiplicative radar speckle, frozen visual features fail to separate spectrally distinct classes, such as wetland versus grass or cropland. Tuning only the language layers, or increasing input resolution, leaves this visual representation fixed (\cref{tab:rendering-design,tab:training-ablation}). We therefore freeze the RGB patch projection and pretrained weights, while we train LoRA modules \citep{lora2022} across attention and feedforward projections, spanning both visual blocks and language layers. During supervised fine-tuning (SFT), we minimize autoregressive cross-entropy over $y=\operatorname{ser}(L^\star,E^\star)$. Visual LoRA parameters employ a lower learning rate, so they adapt mid-level spectral and backscatter features while preserving pretrained visual representations of natural images.

\subsection{Evidence Consistency through EC-DPO}
\label{sec:dpo}

\begin{figure*}[!t]
\centering
\includegraphics[width=0.96\textwidth, trim=23pt 0pt 0pt 0pt, clip]{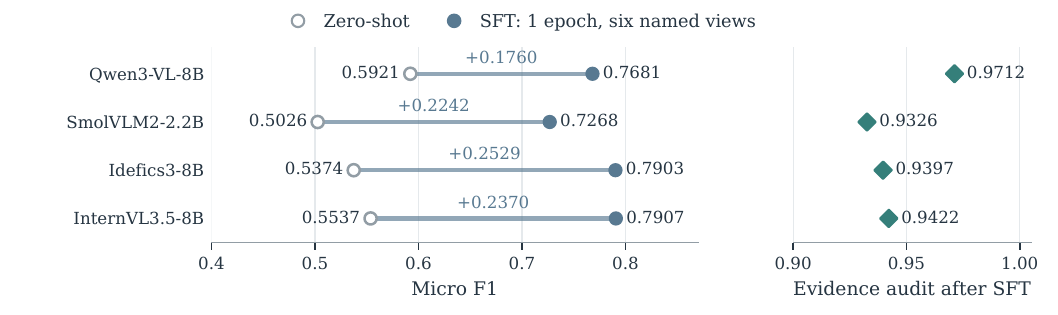}
\caption{Cross-architecture transfer under 1-epoch language LoRA ($n=3{,}234$): $+0.1760$ to $+0.2529$ Micro F1, $>93\%$ audit.}
\label{fig:arch-transfer}
\end{figure*}

After SFT, the reference policy $\pi_{\mathrm{ref}}$ follows the required format. However, $\mathbb{I}_{\mathrm{cue}}(\hat{L},\hat{E})\in\{0,1\}$ checks whether predicted classes in $\hat{L}$ match lexicon phrases in $\hat{E}$, so it ignores omitted classes $c\in L^\star\setminus\hat{L}$. At decision boundaries, the policy may omit a true class, even when the second line retains the optical or radar cues that support it. Unaligned cue phrases can also appear without a matching class. Standard SFT does not contrast complete responses against alternatives where labels and evidence disagree.

From the training pool, we construct preference triples $(x(s),\chosen,\rejected)$ in~\eqref{eq:ecdpo-pair} that align labels with evidence. The chosen response $\chosen=\operatorname{ser}(L^\star,E^\star)$ pairs $L^\star$ with $E^\star$, whereas the rejected response $\rejected=\operatorname{ser}(L_-,E_-)$ removes one true class $c\in L^\star$ while keeping the supporting evidence text unchanged:
\begin{equation}
\label{eq:ecdpo-pair}
\begin{aligned}
\chosen &= \operatorname{ser}(L^\star, E^\star), \quad
\rejected = \operatorname{ser}(L_-, E_-), \\
L_- &= L^\star \setminus \{c\}, \qquad
E_- = E^\star, \\
\text{subject to}\;\; & c \in L^\star, \quad \exists\, q \in T_c^\star \;\text{occurring in}\; E^\star.
\end{aligned}
\end{equation}

Requiring an exclusive cue $q\in T_c^\star$ in $E^\star$ prevents attributing retained evidence to remaining classes in $L_-$. Both responses share identical inputs and evidence, differing only in whether class $c$ is declared. Let $\bar{\ell}_\pi(y\mid x(s))=\frac{1}{|y|}\log\pi(y\mid x(s))$ denote the length-normalized sequence log-likelihood over $|y|$ tokens. Initializing $\pi_\theta$ from the SFT checkpoint, we optimize the DPO objective \citep{dpo2023} in~\eqref{eq:dpo}, anchored by the frozen reference $\pi_{\mathrm{ref}}=\operatorname{sg}(\pi_{\mathrm{SFT}})$ where $\operatorname{sg}(\cdot)$ denotes stop-gradient:
\begin{equation}
\label{eq:dpo}
\begin{aligned}
\mathcal{L}_{\text{EC-DPO}}(\theta)
&= -\log \sigma(\beta \Delta), \\
\Delta
&= \bigl[\bar{\ell}_{\pi_\theta}(\chosen\mid x(s)) - \bar{\ell}_{\pi_{\mathrm{ref}}}(\chosen\mid x(s))\bigr] \\
&\quad - \bigl[\bar{\ell}_{\pi_\theta}(\rejected\mid x(s)) - \bar{\ell}_{\pi_{\mathrm{ref}}}(\rejected\mid x(s))\bigr],
\end{aligned}
\end{equation}
where $\sigma(\cdot)$ is the logistic sigmoid, $\beta>0$ scales the reward penalty, and $\beta \Delta$ is the implicit reward margin. To examine how shared evidence acts prior to length normalization, autoregressive factorization decomposes the raw log-probability ratio into two terms:
\begin{equation*}
\begin{aligned}
\log\frac{\pi_\theta(\chosen\mid x(s))}{\pi_\theta(\rejected\mid x(s))}
&= \log\frac{\pi_\theta(L^\star\mid x(s))}{\pi_\theta(L_-\mid x(s))} \\
&\quad + \log\frac{\pi_\theta(E^\star\mid x(s),L^\star)}{\pi_\theta(E^\star\mid x(s),L_-)}.
\end{aligned}
\end{equation*}
The first term (boundary label recovery) penalizes dropping a class $c\in L^\star$ when its sensor cues are present, which improves recall on ambiguous boundary classes. The second term (conditional evidence coupling) arises since $E_-=E^\star$ is shared. Maximizing this ratio increases the likelihood of generating $E^\star$ when given the complete prefix $L^\star$, while reducing it under the incomplete prefix $L_-=L^\star\setminus\{c\}$. Thus, the model learns to emit exclusive cues $q\in T_c^\star$ only when class $c$ is declared, which reduces unsupported cues while updating both visual and language LoRA parameters.

\section{Experiments}

\subsection{Experimental Setup}
\label{sec:exp-setup}

On BigEarthNet-v2 \citep{bigearthnetmm2021,reben2024}, we merge 18 categories into six classes and use 12,900 official-train patches for training, 3,234 official-test patches for configuration selection, and 3,234 disjoint official-validation patches for final evaluation (Appendix~\ref{app:train}). Six named $224\times224$ views yield 384 visual tokens. Alongside Qwen3-VL-8B-Instruct \citep{qwen3vl2025}, we test five general-purpose VLMs: Qwen2.5-VL-7B \citep{qwen25vl2025}, AdaptLLM-RS-3B \citep{adaptllm2025}, InternVL3.5-8B \citep{internvl35_2025}, Idefics3-8B \citep{idefics3_2024}, and SmolVLM2-2.2B \citep{smolvlm2025}. We also compare four linear probes on TerraFM \citep{terrafm2025}, CROMA \citep{croma2023}, ResNet50 \citep{resnet2016}, and frozen Qwen3-VL, plus four remote-sensing VLMs: SkySenseGPT \citep{skysensegpt2024}, GeoChat \citep{geochat2024}, EarthDial \citep{earthdial2024}, and EarthMind \citep{earthmind2025} (Appendix~\ref{app:baseline-provenance}).

During SFT, we freeze the backbone and add rank-16 LoRA ($\alpha=32$) across language and visual blocks. This adds 51.35M parameters, splitting 43.65M language and 7.70M visual weights in a 196\,MiB checkpoint. We set the language learning rate to $3\times10^{-5}$, with a $0.3\times$ visual scale. EC-DPO trains from SFT on 2,000 pairs ($\beta=0.02$, $5\times10^{-7}$ learning rate). For transfer, we evaluate Sen1Floods11 \citep{sen1floods11} (674 train, 178 test claims) and BigEarthNet.txt \citep{bigearthnettxt2026} (11,377 train, 970 test captions).

\begin{table*}[!t]
\centering
\caption{Multi-view rendering design space across 20 training configurations on 3,234 validation patches (rank-16 language LoRA, Qwen3-VL native grid processor). Canonical reference A4 (\texttt{G1-d}: six named views, $224\times224$, $2\text{--}98\%$ stretch) serves as the shared reference for Blocks A--D. Bold and underline denote the best and second-best results per column.}
\label{tab:rendering-design}
\footnotesize
\setlength{\tabcolsep}{4.5pt}
\renewcommand{\arraystretch}{1.05}
\begin{tabular*}{\textwidth}{@{\extracolsep{\fill}}lllrrrr@{}}
\toprule
Block & Arm ID & Rendering configuration / varied factor & Micro F1 & Macro F1 & Exact match & Ev.\ audit \\
\midrule
\multicolumn{7}{l}{\textit{Block A: View-count ladder (progressive spectral and radar expansion at $224\times224$)}}\\
A1 & G1-a & True color + SAR (2 views, 128 visual tokens) & 0.7880 & 0.6939 & 0.3426 & 0.9703 \\
A2 & G1-b & True + False color + SWIR + SAR (4 views, 256 tokens) & 0.7920 & 0.6857 & 0.3485 & \textbf{0.9774} \\
A3 & G1-c & True + False color + SWIR + NDVI + SAR (5 views, 320 tokens) & 0.7885 & 0.6921 & 0.3469 & 0.9746 \\
A4 & G1-d (ref) & Full six named views (+ NDBI, 6 views, 384 tokens, $2\text{--}98\%$ stretch) & 0.7939 & \textbf{0.6995} & 0.3578 & 0.9712 \\
A5 & G1-e & Seven views (+ NDWI water index, 7 views, 448 tokens) & 0.7877 & 0.6963 & 0.3349 & 0.9657 \\
\midrule
\multicolumn{7}{l}{\textit{Block B: Radiometric contrast stretch (6 views, 384 tokens, vs.\ reference A4 at $2\text{--}98\%$ clipping)}}\\
B1 & G3-a & Full dynamic range ($0\text{--}100\%$) & \textbf{0.7957} & 0.6942 & \textbf{0.3726} & 0.9728 \\
B2 & G3-b & Wide clipping ($1\text{--}99\%$) & 0.7949 & \underline{0.6994} & 0.3575 & 0.9694 \\
B3 & G3-c & Tight clipping ($5\text{--}95\%$) & 0.7912 & 0.6941 & 0.3491 & 0.9759 \\
B4 & G3-d & Gaussian standardization ($\mu \pm 2\sigma$) & \underline{0.7954} & 0.6990 & \underline{0.3639} & 0.9716 \\
\midrule
\multicolumn{7}{l}{\textit{Block C: Spectral index colormaps and composite band triples (6 views, 384 tokens)}}\\
C1 & G4-a & Narrow diverging colormap ($[-0.5, +0.5]$) & 0.7870 & 0.6895 & 0.3488 & 0.9709 \\
C2 & G4-b & Grayscale index rendering & 0.7879 & 0.6815 & 0.3423 & 0.9731 \\
C3 & G4-c & Sign-thresholded bipolar rendering & 0.7889 & 0.6908 & 0.3377 & 0.9666 \\
C4 & G5-a & Alternative SWIR (B12, B8A, B04) & 0.7899 & 0.6911 & 0.3451 & 0.9709 \\
C5 & G5-b & Dual-SWIR narrow-NIR (B11, B12, B8A) & 0.7937 & 0.6919 & 0.3547 & 0.9706 \\
C6 & G5-c & SWIR-substituted false color (B11, B08, B04) & 0.7885 & 0.6869 & 0.3448 & 0.9725 \\
\midrule
\multicolumn{7}{l}{\textit{Block D: View naming and sequence ordering (6 identical pixel views, vs.\ canonical reference A4)}}\\
D1 & G6-a & Generic indices (\texttt{Image 1..6}) & 0.7884 & 0.6896 & 0.3541 & 0.9728 \\
D2 & G6-b & Coarse family tags (\texttt{MSI} vs.\ \texttt{SAR}) & 0.7913 & 0.6984 & 0.3503 & 0.9725 \\
D3 & G6-d & Explicit physical names with shuffled order & 0.7870 & 0.6897 & 0.3398 & 0.9734 \\
\midrule
\multicolumn{7}{l}{\textit{Block E: Spatial resolution and visual token budget (6 views, 3,072 max sequence length)}}\\
E1 & G6-e-ref & Standard $224\times224$ ($64$ tokens/view, $384$ visual tokens) & 0.7930 & 0.6958 & 0.3575 & 0.9734 \\
E2 & G6-e & High-res $448\times448$ ($196$ tokens/view, $1{,}176$ visual tokens) & \underline{0.7954} & 0.6933 & 0.3593 & \underline{0.9771} \\
\bottomrule
\end{tabular*}
\end{table*}

\subsection{Main Results and Architecture Transfer}
\label{sec:main-results}

\Cref{tab:main} compares our framework across four tiers. Tier~1 linear probes give frozen-encoder references (11,651 training patches), while Tier~2 remote-sensing VLMs suffer zero-shot drops from RGB-only interfaces and land-cover taxonomy mismatch.

Tier~3 and \cref{fig:arch-transfer} show that our named six-view interface transfers cleanly across four VLM families. Language LoRA lifts Micro F1 by $+0.1760$ to $+0.2529$ across Qwen3-VL-8B, SmolVLM2-2.2B, Idefics3-8B, and InternVL3.5-8B, while Qwen2.5-VL-7B and AdaptLLM-RS-3B reach 0.8026 and 0.7917. All six backbones achieve 100.0\% format validity, and they exceed 91.5\% lexical audit compliance. In Tier~4, switching Qwen3-VL-8B from the generic wrapper at 0.7681 to its native multi-image processor recovers $+0.0258$, reaching 0.7939 in \cref{tab:rendering-design} (\texttt{G1-d}). Joint visual and language SFT then reaches 0.8242 Micro F1 and 0.4288 Exact Match. Adding EC-DPO lifts Qwen3-VL-8B to \textbf{0.8526} Micro F1, \textbf{0.7783} Macro F1, and \textbf{0.4720} Exact Match. With lightweight adapters, it matches or exceeds frozen-encoder linear probes such as CROMA (0.8334) and TerraFM (0.8442), while generating verifiable multisensor rationales.

\subsection{Multi-View Rendering Design Space and Precision--Recall Trade-Offs}
\label{sec:rendering-design}

\begin{figure*}[!t]
\centering
\begin{subfigure}[t]{0.47\textwidth}
\centering
\includegraphics[width=\linewidth]{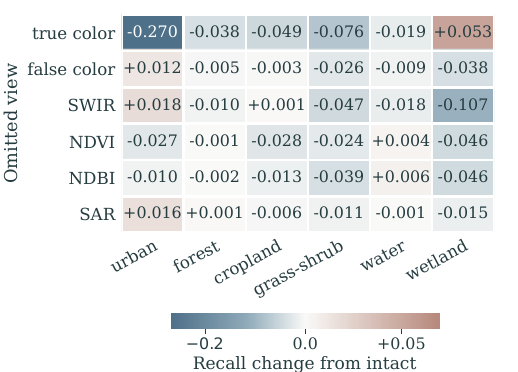}
\caption{Per-class recall drops.}
\label{fig:lovo-heat}
\end{subfigure}
\hfill
\begin{subfigure}[t]{0.47\textwidth}
\centering
\includegraphics[width=\linewidth]{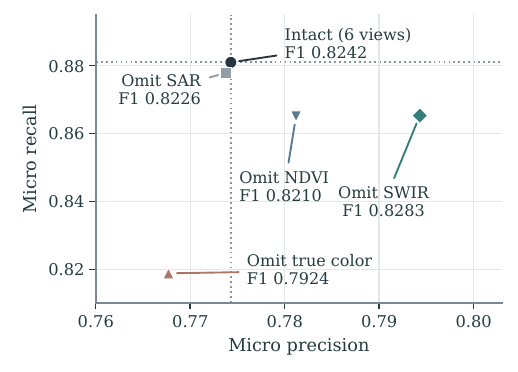}
\caption{Precision--recall trade-off (representative arms).}
\label{fig:lovo-pr-scatter}
\end{subfigure}
\caption{Multi-view reliance under LOVO omission ($3{,}234$ patches). False-color and NDBI omissions sit close to Intact and are omitted in (b) for legibility; full rows appear in Supplementary Table~\ref{tab:lovo}.}
\label{fig:lovo-pr}
\end{figure*}

\begin{table*}[!t]
\centering
\caption{Three-seed LoRA placement with parameter-matched $r=19$ language LoRA (Part~a; Supplementary Table~\ref{tab:per-seed-lora}) and EC-DPO controls ($n=3{,}234$, Part~b). Bold/underline: best/second-best per part.}
\label{tab:training-ablation}
\footnotesize
\setlength{\tabcolsep}{4.0pt}
\begin{tabular*}{\textwidth}{@{\extracolsep{\fill}}lrrrrrrr@{}}
\toprule
Adaptation Stage \& Scope & Trainable & Micro F1 & Macro F1 & Exact & Wet.\ Rec. & Wat.\ Rec. & Ev.\ audit \\
\midrule
\multicolumn{8}{l}{\textit{Part (a): Multi-host 3-seed reproducibility \& parameter-matched LoRA placement suite (seeds 42 / 123 / 2026)}}\\
Language-only LoRA SFT ($r=16$) & 43.65M & \pmval{0.7998}{0.0057} & \pmval{0.7195}{0.0054} & \pmval{0.3654}{0.0054} & \pmval{0.6590}{0.0044} & \pmval{\mathbf{0.9133}}{0.0069} & 0.9714 \\
Language-only LoRA SFT ($r=19$, param-matched) & 51.84M & \pmval{0.8008}{0.0037} & \pmval{0.7203}{0.0034} & \pmval{0.3665}{0.0073} & \pmval{0.6718}{0.0202} & \pmval{0.9118}{0.0096} & 0.9718 \\
\textbf{Joint V+L LoRA SFT ($r=16$, Ours)} & 51.35M & \pmval{\mathbf{0.8309}}{0.0053} & \pmval{\mathbf{0.7416}}{0.0071} & \pmval{\mathbf{0.3985}}{0.0165} & \pmval{\mathbf{0.7812}}{0.0117} & \pmval{0.8988}{0.0040} & \textbf{0.9741} \\
\textit{$\Delta$ (Joint V+L vs.\ Param-matched Lang $r=19$)} & $-0.49\text{M}$ & $\mathbf{+0.0301}$ & $\mathbf{+0.0213}$ & $\mathbf{+0.0320}$ & $\mathbf{+0.1094}$ & $-0.0130$ & $+0.0023$ \\
\midrule
\multicolumn{8}{l}{\textit{Part (b): Primary confirmatory checkpoint \& post-training boundary preference lineage ($n=3{,}234$)}}\\
Joint V+L SFT (init.) & 51.35M & 0.8242 & 0.7561 & 0.4288 & 0.7481 & 0.9068 & \underline{0.9753} \\
+ Chosen-only SFT control & 51.35M & 0.8337 & 0.7516 & 0.4112 & 0.7405 & 0.9052 & 0.9712 \\
+ Label-only DPO (class line only, no evidence) & 51.35M & 0.7379 & 0.6792 & 0.2035 & 0.6183 & 0.8756 & -- \\
+ Coupled cue removal DPO ($E_- = E^\star \setminus T_c^\star$) & 51.35M & \underline{0.8363} & \underline{0.7579} & \underline{0.4344} & \underline{0.7641} & \underline{0.9078} & 0.9746 \\
\textbf{+ EC-DPO preference (Ours, shared $E_- = E^\star$)} & 51.35M & \textbf{0.8526} & \textbf{0.7783} & \textbf{0.4720} & \textbf{0.8010} & \textbf{0.9121} & \textbf{0.9829} \\
\textit{$\Delta$ (EC-DPO vs.\ SFT init.)} & $0\text{M}$ & $+0.0284$ & $+0.0222$ & $\mathbf{+0.0432}$ & $\mathbf{+0.0529}$ & $+0.0053$ & $+0.0076$ \\
\textit{$\Delta$ (EC-DPO vs.\ Chosen-only)} & $0\text{M}$ & $+0.0189$ & $+0.0267$ & $\mathbf{+0.0608}$ & $\mathbf{+0.0605}$ & $+0.0069$ & $+0.0117$ \\
\bottomrule
\end{tabular*}
\end{table*}

\begin{table}[!t]
\centering
\captionsetup{skip=4pt}
\caption{Counterfactual interventions on clear-sky ($n=800$) and \texttt{BENv2-Cloud} ($n=1{,}000$) patches. $\Delta$Pred measures label-set changes from intact inference. Donor swaps retain source labels; row~(f) replaces false color, SWIR, NDVI, and NDBI, keeping true color and SAR. Row~(f), clear sky: matched seed-42 intact F1 $=0.8162$ (Appendix~\ref{app:image-dependence}).}
\label{tab:image-dependence}
\footnotesize
\setlength{\tabcolsep}{2.0pt}
\renewcommand{\arraystretch}{0.96}
\begin{tabular*}{\columnwidth}{@{\extracolsep{\fill}}lcccc@{}}
\toprule
& \multicolumn{2}{c}{Clear-Sky ($800$)} & \multicolumn{2}{c}{BENv2-Cloud ($1{,}000$)} \\
\cmidrule(lr){2-3} \cmidrule(lr){4-5}
Intervention & Micro F1 & $\Delta$Pred & Micro F1 & $\Delta$Pred \\
\midrule
(a) Intact (5 opt + SAR) & 0.8180 & 0.00\% & \textbf{0.7654} & 0.00\% \\
(b) MSI only (no SAR) & 0.8159 & 11.00\% & 0.7613 & \textbf{23.00\%} \\
(c) SAR only (no optical) & 0.1734 & 88.88\% & 0.3909 & 81.30\% \\
(d) Optical donor + SAR & 0.4852 & 93.25\% & 0.3746 & 93.00\% \\
(e) Optical + SAR donor & 0.8186 & 5.75\% & 0.7558 & \textbf{22.80\%} \\
(f) + 4 spectral donors & 0.7287 & \textbf{62.88\%} & 0.6960 & \textbf{53.10\%} \\
(g) Blind text prior & 0.5741 & 97.00\% & 0.5734 & 96.10\% \\
\bottomrule
\end{tabular*}
\end{table}

\begin{table*}[!t]
\centering
\caption{Cross-task transfer on (a) Sen1Floods11 flood verification (four views: true color, SWIR, NDWI, SAR) and (b) BigEarthNet.txt dense captioning (Concept F1, SacreBLEU, ROUGE-L; six views) with joint LoRA. Bold and underline mark best and second-best values.}
\label{tab:transfer}
\footnotesize
\setlength{\tabcolsep}{2.5pt}
\begin{tabular*}{\textwidth}{@{\extracolsep{\fill}}lrrrrlrrr@{}}
\toprule
\multicolumn{5}{c}{\textbf{(a) Sen1Floods11 Flood Verification ($178$ claims, $89$ patches)}} & \multicolumn{4}{c}{\textbf{(b) BigEarthNet.txt Dense Captioning ($970$ captions)}} \\
\cmidrule(lr){1-5} \cmidrule(lr){6-9}
Setting & Flood F1 & Non-Flood & Macro & Acc. & Setting & Concept F1 & BLEU & ROUGE-L \\
\midrule
Qwen3-VL Zero-shot (MSI+SAR) & 0.5294 & 0.5699 & 0.5497 & 0.5506 & Base Qwen3-VL Zero-shot & 0.0148 & 0.81 & 0.1177 \\
SFT without images (blind) & 0.0000 & 0.7724 & 0.3862 & 0.6292 & SFT without images (blind) & 0.2753 & 31.88 & 0.4421 \\
SFT SAR-only (radar) & 0.6604 & \textbf{0.8560} & \underline{0.7582} & \underline{0.7978} & SFT Language-only LoRA (6 views) & \underline{0.4270} & \underline{38.13} & \underline{0.5090} \\
SFT MSI-only (3 views) & \underline{0.7072} & 0.6971 & 0.7022 & 0.7022 & \textbf{SFT Joint V+L LoRA (6 views)} & \textbf{0.5654} & \textbf{42.07} & \textbf{0.5337} \\
\textbf{SFT MSI + SAR (4 views)} & \textbf{0.7714} & \underline{0.8519} & \textbf{0.8116} & \textbf{0.8202} & \textit{$\Delta$ (Joint V+L vs.\ Lang-only)} & $\mathbf{+0.1384}$ & $\mathbf{+3.94}$ & $\mathbf{+0.0247}$ \\
\bottomrule
\end{tabular*}
\end{table*}

\Cref{tab:rendering-design} tests 20 rendering configurations across five axes, revealing three principles. In Block~A, (i) \textbf{shared spectral bands keep accuracy stable across view counts and stretches}. True color plus SAR reaches 0.7880 Micro F1, while six views reach 0.7939. Adding a seventh normalized difference water index (NDWI) view drops Micro F1 to 0.7877, adding tokens while reusing Green B03 and NIR B08. Blocks~B and~C stay stable across stretches (0.7912--0.7957) and SWIR mixes with B12 and narrow-NIR B8A (0.7885--0.7937), whereas grayscale indices drop Macro F1 by $-0.0180$ by erasing sign polarity. In Block~E, $448\times448$ inputs triple visual tokens, changing Micro F1 by just $+0.0024$. Increasing input resolution yields only a small gain with the frozen encoder. In Block~D, (ii) \textbf{named views and fixed ordering govern alignment}. Fixed slot order already yields 0.9728 lexical audit on generic indices (\texttt{D1}). Physical names add semantic alignment gains over \texttt{D1} ($+0.0055$ Micro F1, $+0.0099$ Macro F1), while shuffling view order drops Exact Match by $-0.0180$ (Appendix~\ref{app:inputablation}).

At the class level, (iii) \textbf{derived views govern minority precision and recall}. Under leave-one-view-out (LOVO) inference (Supplementary Table~\ref{tab:lovo}, \cref{fig:lovo-pr}), removing true color hurts most, cutting Micro F1 by $-0.0318$ and urban recall by $-26.99$ percentage points (pp). True color supplies the exclusive B02 band and shares B03 with false color; Red B04 and NIR B08 span four views. Omitting SWIR nudges Micro F1 up $+0.0041$ to 0.8283 by reducing predictions from 2.489 to 2.383 labels per patch. This cuts wetland false alarms (378 to 230) and lifts precision (0.7743 to 0.7943), yet it drops \textbf{wetland recall by $-10.69$ pp} ($74.81\%$ to $64.12\%$), where wetland forms $1.9\%$ of labels. Omitting NDVI, false color, or NDBI also cuts wetland recall ($-3.82$ to $-4.58$ pp), confirming the importance of derived views.

\begin{figure}[!t]
\centering
\includegraphics[width=\linewidth]{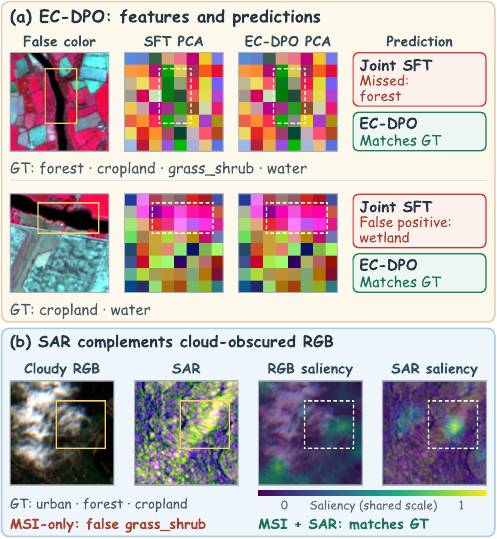}
\caption{Three qualitative cases. (a)~Compared with Joint SFT, EC-DPO recovers \texttt{forest} (top) and removes false \texttt{wetland} (bottom). The $8\times8$ false-color tokens share a PCA basis per scene; colors do not denote classes. (b)~On \texttt{BENv2-Cloud}, MSI+SAR removes MSI-only's false \texttt{grass\_shrub}; RGB/SAR gradient saliency shares one normalized scale. GT: ground truth. Boxes mark corresponding regions.}
\label{fig:qualitative-dpo}
\end{figure}

\subsection{Joint Vision-Language Tuning and EC-DPO}
\label{sec:dpo-and-human}

\Cref{tab:training-ablation}(a) tests whether language tuning suffices for rendered views. Raising language rank ($r=16$ to $r=19$, 51.84M parameters) shifts Micro F1 by just $+0.0010$ to 0.8008. Moving 7.70M parameters into visual blocks (Joint V+L LoRA, 51.35M) outperforms matched language LoRA by $\mathbf{+0.0301}$ Micro F1, $\mathbf{+0.0320}$ Exact Match, and $\mathbf{+10.94}$ pp wetland recall ($67.18\%$ to $78.12\%$). Alongside the $+0.0024$ gain at $448\times448$, this shows visual LoRA separates false-color, index, and speckle patterns.

In \cref{tab:training-ablation}(b), EC-DPO reaches \textbf{0.8526} Micro F1 and \textbf{0.4720} Exact Match, gaining $+0.0432$ over SFT (paired bootstrap, $p < 0.001$), with wetland recall up $+5.29$ pp to \textbf{0.8010}. Three controls isolate the term in \cref{sec:dpo}. Chosen-only SFT on 2,000 positive responses omits the boundary contrast, cutting Exact Match to 0.4112. Label-only DPO drops the evidence line, removing the anchor and dropping Exact Match to 0.2035. Stripping $T_c^\star$ from $E_-$ breaks the shared cue condition, reaching 0.8363 Micro F1 and 0.4344 Exact Match. Across 100 scenes and 400 outputs in double-blind audit (Supplementary Table~\ref{tab:human-audit}), EC-DPO boosts consistency from $81.5\%$ to \textbf{95.5\%} and cuts hallucinations from $18.5\%$ to \textbf{4.5\%} (\cref{fig:qualitative-dpo}a).

\subsection{Counterfactual Sensor Interventions and All-Weather Resilience}
\label{sec:sensor-interventions}

\Cref{tab:image-dependence} explains why single-view LOVO (Supplementary Table~\ref{tab:lovo}) barely shifts Micro F1. Swapping all four derived views with donors while keeping true color and SAR intact (Row~f) alters \textbf{62.88\%} of predictions, cutting Micro F1 by $\mathbf{-0.0875}$ (0.8162 to 0.7287) and Exact Match from 0.4088 to 0.2213. Single-view omission leaves shared bands like NIR B08 intact, so micro-averaging masks wetland drops. Replacing four derived views breaks this redundancy, dropping wetland recall by $-39.40$ pp and water recall by $-36.08$ pp.

On BENv2-Cloud, clear and cloudy splits reveal adaptive radar reliance. Under clear skies, optical views dominate, and SAR swapping (Row~e) alters just $5.75\%$ of predictions. Under cloud cover, SAR swapping and removal alter \textbf{22.80\%} and \textbf{23.00\%} of predictions, while derived-view swaps (Row~f) cut Micro F1 from 0.7654 to 0.6960 and flip \textbf{53.10\%} of labels. Microwave penetration lets SAR-only inference (Row~c) retain 0.6463 water F1 and 0.6328 forest F1 (\cref{fig:qualitative-dpo}b).

\subsection{Cross-Task Generalization: Flood Verification and Dense Captioning}
\label{sec:transfer}

\Cref{tab:transfer} evaluates adaptation on flood verification and dense captioning. On Sen1Floods11, joint MSI+SAR input improves flood recognition over either sensor alone, reaching \textbf{0.7714} Flood F1 ($+0.0642$ over MSI-only) and \textbf{82.02\%} accuracy. On BigEarthNet.txt, visual adaptation improves land-cover descriptions using the same six rendered views: joint V+L LoRA reaches \textbf{0.5654} Concept F1 and \textbf{42.07} SacreBLEU, gains of $\mathbf{+0.1384}$ and $\mathbf{+3.94}$ over language-only tuning.

\section{Conclusion}
\label{sec:conclusion}

We introduced MVSA, adapting general-purpose VLMs to MSI and SAR without modifying backbones. Named six-view rendering resolves channel ambiguity, while joint LoRA adapts visual and language blocks to spectral and radar signals. In post-training, EC-DPO aligns labels with evidence, raising boundary recall and pruning unsupported cues. Ablations and cloud interventions confirm that shared bands protect dominant classes, while derived views and SAR sustain wetland recall and occluded inference.

\textbf{Limitations.} Rendering six views expands each patch to 384 visual tokens, increasing prefill compute and memory over single-image inputs. Per-patch $2\text{--}98\%$ stretching also maps observations to 8-bit RGB, removing calibrated reflectance and SAR backscatter units. Merging 18 categories into six classes trades fine-grained taxonomy for cross-sensor stability.

\clearpage

\section*{LLM Usage Statement}
In this work, the generative AI tool Codex was used to polish the manuscript text and format explanatory illustrations. All AI-assisted materials were reviewed and edited by the authors. The authors take full responsibility for the research design, scientific claims, experimental results, citations, and the final manuscript, including all text, claims, and artifacts produced with AI assistance.

\clearpage
\setcounter{figure}{0}
\setcounter{table}{0}
\setcounter{secnumdepth}{2}
\renewcommand{\theHfigure}{supp.\arabic{figure}}
\renewcommand{\theHtable}{supp.\arabic{table}}
\appendix
\providecommand{\pmval}[2]{\begin{tabular}[c]{@{}c@{}}$#1$\\[-2pt]\scriptsize$\pm#2$\end{tabular}}

\section{Supplemental Material Overview}

This supplement documents dataset splits, hyperparameters, and label taxonomy rules. It also details lexicon auditing dictionaries, sensor-intervention controls, human expert scoring, cross-task evaluation setups, baseline provenance, and extended qualitative results corresponding to \mbox{Main Tables~\ref{tab:main}--\ref{tab:transfer}}.

\subsection{Dataset Splits and Optimization Setup}
\label{app:train}

All multisensor experiments use aligned Sentinel-2 MSI and Sentinel-1 SAR scenes from BigEarthNet-v2, discarding patches without a mapped land-cover class. Training uses 12,900 patches from the official training split, and configuration checks use 3,234 official test patches. Final evaluation runs on 3,234 disjoint validation patches (seed 42).

Table~\ref{tab:app-settings} lists all optimization settings. During supervised fine-tuning (SFT), pretrained Qwen3-VL-8B weights stay frozen. We attach rank-16 LoRA adapters ($\alpha=32$) across the decoder and visual blocks, targeting query, key, value, output, and feedforward layers. Visual adapters scale the $3\times10^{-5}$ language rate by $0.3\times$, yielding 51.35M trainable weights (43.65M language, 7.70M visual) in 196\,MiB.

\begin{table*}[!t]
\centering
\caption{Controlled three-seed LoRA placement breakdown on the 3,234-patch validation split corresponding to Main Table~\ref{tab:training-ablation}(a). Parameter-matched language-only LoRA ($r=19$) matches the trainable parameter budget of joint vision--language adaptation ($51.84\text{M}$ vs.\ $51.35\text{M}$).}
\label{tab:per-seed-lora}
\footnotesize
\setlength{\tabcolsep}{3.5pt}
\begin{tabular*}{\textwidth}{@{\extracolsep{\fill}}llrrrrr@{}}
\toprule
Adaptation Scope & Seed & Params & Micro F1 & Macro F1 & Exact match & Wet.\ rec. \\
\midrule
Lang-only LoRA ($r=16$) & 42 & 43.65M & 0.7951 & 0.7144 & 0.3661 & 0.6565 \\
Lang-only LoRA ($r=16$) & 123 & 43.65M & 0.8062 & 0.7252 & 0.3596 & 0.6565 \\
Lang-only LoRA ($r=16$) & 2026 & 43.65M & 0.7981 & 0.7188 & 0.3704 & 0.6641 \\
\textit{Mean $\pm$ SD ($r=16$ Lang)} & 3 seeds & 43.65M & \pmval{0.7998}{0.0057} & \pmval{0.7195}{0.0054} & \pmval{0.3654}{0.0054} & \pmval{0.6590}{0.0044} \\
\midrule
Lang-only LoRA ($r=19$, matched) & 42 & 51.84M & 0.7967 & 0.7170 & 0.3670 & 0.6565 \\
Lang-only LoRA ($r=19$, matched) & 123 & 51.84M & 0.8040 & 0.7202 & 0.3590 & 0.6641 \\
Lang-only LoRA ($r=19$, matched) & 2026 & 51.84M & 0.8018 & 0.7238 & 0.3735 & 0.6947 \\
\textit{Mean $\pm$ SD ($r=19$ Lang)} & 3 seeds & 51.84M & \pmval{0.8008}{0.0037} & \pmval{0.7203}{0.0034} & \pmval{0.3665}{0.0073} & \pmval{0.6718}{0.0202} \\
\midrule
Joint V+L LoRA ($r=16$, Ours) & 42 & 51.35M & 0.8297 & 0.7417 & 0.3980 & 0.7710 \\
Joint V+L LoRA ($r=16$, Ours) & 123 & 51.35M & 0.8367 & 0.7487 & 0.4153 & 0.7786 \\
Joint V+L LoRA ($r=16$, Ours) & 2026 & 51.35M & 0.8263 & 0.7345 & 0.3822 & 0.7939 \\
\textit{Mean $\pm$ SD (Joint V+L)} & 3 seeds & 51.35M & \pmval{0.8309}{0.0053} & \pmval{0.7416}{0.0071} & \pmval{0.3985}{0.0165} & \pmval{0.7812}{0.0117} \\
\bottomrule
\end{tabular*}
\end{table*}

\begin{table}[t]
\centering
\caption{Summary of training and validation hyperparameters for the main Qwen3-VL-8B-Instruct pipeline.}
\label{tab:app-settings}
\small
\setlength{\tabcolsep}{2.5pt}
\begin{tabularx}{\columnwidth}{@{}P{0.23\columnwidth}Y@{}}
\toprule
Item & Configuration \\
\midrule
\multicolumn{2}{@{}l}{\textit{Data and Evaluation Splits}} \\
Classes (6) & \texttt{urban}, \texttt{forest}, \texttt{cropland}, \texttt{grass\_shrub}, \texttt{water}, \texttt{wetland} \\
Splits & 12,900 train / 3,234 config (test) / 3,234 validation \\
Probes (Tier~1) & 11,651 train / 1,249 threshold / 3,234 val.\ (ResNet50: 90\%/10\% split, 11,610 / 1,290. Mean-pooled frozen features + per-class F1 thresholds) \\
Input format & Six $224\times224$ named RGB-encoded views (384 visual tokens) \\
\midrule
\multicolumn{2}{@{}l}{\textit{Optimization Schedule}} \\
Joint SFT & AdamW, $r=16$ ($\alpha=32$, dropout $0.05$), batch size 1, grad.\ accum.\ 8, 2 epochs, LR $3\times10^{-5}$ ($0.3\times$ visual), max length 1,536, grad.\ checkpointing \\
EC-DPO & Frozen SFT init.\ reference, 2,000 pairs, $\beta=0.02$, 1 epoch, LR $5\times10^{-7}$, max length 1,536, grad.\ clip 1.0 \\
Decoding & Deterministic greedy decoding (\texttt{do\_sample=false}, \texttt{max\_new\_tokens=96}) \\
\bottomrule
\end{tabularx}
\end{table}

\subsection{Label Taxonomy and Sampling}
\label{app:taxonomy}

BigEarthNet-v2 defines 19 land-cover classes. Table~\ref{tab:label-map} merges 18 mapped categories into six unified target labels. We exclude pure coastal sand patches (\textit{beaches, dunes, sands}) lacking another mapped category. On cloud- and snow-free patches, the 12,900-patch train split and 3,234-patch configuration split use greedy class-quota sampling to preserve rare classes. Positive counts in training and configuration are \texttt{urban} (4,032 / 1,006), \texttt{forest} (8,062 / 2,299), and \texttt{cropland} (7,056 / 1,892). Counts for the remaining classes are \texttt{grass\_shrub} (6,346 / 1,672), \texttt{water} (4,256 / 1,084), and \texttt{wetland} (4,000 / 1,000). By contrast, we draw the 3,234-patch validation split uniformly at random (seed 42) from disjoint scenes. This split keeps natural frequencies: 489 \texttt{urban}, 2,076 \texttt{forest}, 2,022 \texttt{cropland}, 1,575 \texttt{grass\_shrub}, 781 \texttt{water}, and 131 \texttt{wetland} positives ($1.85\%$ of positive labels).

\begin{table}[t]
\centering
\caption{Deterministic mapping from 18 BigEarthNet-v2 classes to the six target categories.}
\label{tab:label-map}
\small
\setlength{\tabcolsep}{2.5pt}
\begin{tabularx}{\columnwidth}{@{}P{0.25\columnwidth}Y@{}}
\toprule
Target Class & Merged BigEarthNet-v2 Categories \\
\midrule
\texttt{urban} & Urban fabric / industrial or commercial units \\
\texttt{forest} & Broad-leaved, coniferous, and mixed forest \\
\texttt{cropland} & Arable land / permanent crops / complex cultivation / agro-forestry / agriculture with natural vegetation \\
\texttt{grass\_shrub} & Pastures / natural grassland and sparse vegetation / transitional woodland/shrub / moors, heathland, and sclerophyllous vegetation \\
\texttt{water} & Inland waters / marine waters \\
\texttt{wetland} & Inland wetlands / coastal wetlands \\
\bottomrule
\end{tabularx}
\end{table}

\subsection{Canonical Class Parsing}
\label{app:alias}

During scoring, a deterministic parser lowercases the \texttt{class:} line. It maps lexical aliases such as \texttt{built-up}, \texttt{woodland}, \texttt{agriculture}, \texttt{grassland}, hyphenated forms (\texttt{grass-shrub}), and plurals to the canonical six classes. If an external zero-shot model omits the \texttt{class:} line, the parser records a format failure. It then falls back to full-text alias matching so all baselines share a common evaluation denominator.

\subsection{Multi-View Rendering Protocol}
\label{app:render}

\begin{table}[t]
\centering
\caption{Six-view rendering specification for the physically named MSI+SAR input interface.}
\label{tab:app-render}
\footnotesize
\setlength{\tabcolsep}{2.0pt}
\begin{tabularx}{\columnwidth}{@{}P{0.18\columnwidth}P{0.25\columnwidth}Y@{}}
\toprule
View Name & Channel Mapping & Physical Content Encoded \\
\midrule
True color & B04, B03, B02 & Visible surface appearance \\
False color & B08, B04, B03 & NIR vegetation contrast \\
SWIR & B11, B08, B04 & Built-up and moisture separation \\
NDVI & Diverging RGB & $(\mathrm{B08}-\mathrm{B04})/(\mathrm{B08}+\mathrm{B04})$ \\
NDBI & Diverging RGB & $(\mathrm{B11}-\mathrm{B08})/(\mathrm{B11}+\mathrm{B08})$ \\
SAR & VV, VH, VV$-$VH & Log backscatter \& polarization \\
\bottomrule
\end{tabularx}
\end{table}

All Sentinel-2 bands resample bilinearly onto the 10\,m B04 grid at $224\times224$ pixels. Under Qwen3-VL's native grid processor, each view yields 64 visual tokens without cropping (384 visual tokens total). Each optical channel and calibrated decibel SAR channel ($\mathrm{VV}$, $\mathrm{VH}$, $\mathrm{VV}-\mathrm{VH}$) clips between its per-patch 2nd and 98th finite percentiles. Clipped values then scale linearly to 8-bit RGB integers. The scalar spectral indices use $\mathrm{NDVI}=(\mathrm{B08}-\mathrm{B04})/(\mathrm{B08}+\mathrm{B04})$ and $\mathrm{NDBI}=(\mathrm{B11}-\mathrm{B08})/(\mathrm{B11}+\mathrm{B08})$. In the seven-view sweep, $\mathrm{NDWI}=(\mathrm{B03}-\mathrm{B08})/(\mathrm{B03}+\mathrm{B08}+10^{-6})$ completes the set. Each index maps over $[-1,1]$, using a zero-centered diverging blue--red ramp (Table~\ref{tab:app-render}). For Main Table~\ref{tab:rendering-design}(E), the $448\times448$ run upsamples the same rendered pixels. Relative to its matched control (\texttt{G6-e-ref}, 0.7930 vs.\ 0.7939 for \texttt{G1-d}), it raises the sequence cap from 1,536 to 3,072 tokens so all visual tokens fit without truncation.

\subsection{Evidence Target Construction}
\label{app:templates}

Targets $E^\star$ use programmatic class cue templates. Each template pairs an optical/SWIR/index clause, assigned to MSI, with a backscatter clause, assigned to SAR. Standardized sensor phrases keep the supervision text consistent. Table~\ref{tab:app-dict} gives each class $c$ a disjoint lexicon $T_c^\star = T_c$, letting coupled controls strip cues via $E_- = E^\star \setminus T_c^\star$.

\begin{table}[t]
\centering
\caption{Double-blind human expert audit across 100 scenes (400 randomized outputs, Cohen's $\kappa = 0.866\text{--}0.895$), alongside the unadapted zero-shot reference.}
\label{tab:human-audit}
\footnotesize
\setlength{\tabcolsep}{3.0pt}
\begin{tabular*}{\columnwidth}{@{\extracolsep{\fill}}lccc@{}}
\toprule
Evaluation Condition ($100$ scenes) & Spec.\ $\uparrow$ & Phys.\ $\uparrow$ & Halluc.\ $\downarrow$ \\
\midrule
\multicolumn{4}{l}{\textit{Unadapted reference ($100$ outputs)}}\\
Zero-shot Qwen3-VL (6 views) & 1.82 & 42.0\% & 58.0\% \\
\midrule
\multicolumn{4}{l}{\textit{Randomized double-blind pool ($400$ outputs)}}\\
Joint LoRA SFT (6 views) & \underline{3.67} & \underline{81.5\%} & \underline{18.5\%} \\
\textbf{Joint LoRA + EC-DPO (6 views)} & \textbf{4.31} & \textbf{95.5\%} & \textbf{4.5\%} \\
Joint LoRA SFT (blind, no images) & 1.27 & 29.0\% & 65.5\% \\
Joint LoRA + EC-DPO (blind) & 1.26 & 25.0\% & 69.0\% \\
\bottomrule
\end{tabular*}
\end{table}

\subsection{Evidence Dictionaries and Lexical Audit}
\label{app:dict}

\begin{table}[t]
\centering
\caption{Audited source and class-cue substring lexicons ($T_c$) used for automated lexical checking $\mathcal{A}(\hat{L},\hat{E})$. For sensor attribution ($\mathbb{I}_{\mathrm{attr}}$), \texttt{*} marks optical-only and SAR-only substrings that cannot cross-pair without a matching source name.}
\label{tab:app-dict}
\scriptsize
\setlength{\tabcolsep}{2.5pt}
\begin{tabularx}{\columnwidth}{@{}P{0.22\columnwidth}Y@{}}
\toprule
Lexicon Group & Substring Entries (Case-Insensitive) \\
\midrule
Optical / MSI sources & \texttt{msi}, \texttt{optical}, \texttt{multispectral}, \texttt{true-color}*, \texttt{true color}*, \texttt{false-color}*, \texttt{false color}*, \texttt{swir}*, \texttt{ndvi}*, \texttt{ndbi}*, \texttt{spectral}*, \texttt{vegetation-sensitive}* \\
SAR sources & \texttt{sar}, \texttt{radar}*, \texttt{backscatter}*, \texttt{double-bounce}*, \texttt{double bounce}*, \texttt{speckle}* \\
\texttt{urban} cues & \texttt{built-up}, \texttt{built up}, \texttt{impervious}, \texttt{ndbi}, \texttt{man-made}, \texttt{urban block}, \texttt{structured radar}, \texttt{angular}, \texttt{compact bright} \\
\texttt{forest} cues & \texttt{tree canopy}, \texttt{canopy}, \texttt{forest canopy}, \texttt{closed canopy}, \texttt{volume scattering}, \texttt{dense continuous}, \texttt{vegetation-rich}, \texttt{rich vegetation} \\
\texttt{cropland} cues & \texttt{field parcel}, \texttt{field-like}, \texttt{agricultural}, \texttt{crop}, \texttt{cultivated}, \texttt{parcel}, \texttt{regular surface}, \texttt{regular texture} \\
\texttt{grass\_shrub} cues & \texttt{grass}, \texttt{shrub}, \texttt{low-to-moderate}, \texttt{low to moderate}, \texttt{herbaceous}, \texttt{pasture}, \texttt{mottled natural}, \texttt{moderate diffuse} \\
\texttt{water} cues & \texttt{open water}, \texttt{dark water}, \texttt{water-like}, \texttt{water-compatible}, \texttt{smooth low-variation}, \texttt{low backscatter}, \texttt{aquatic}, \texttt{water-sensitive} \\
\texttt{wetland} cues & \texttt{wet vegetation}, \texttt{moisture}, \texttt{moisture-sensitive}, \texttt{mixed water}, \texttt{mixed water-vegetation}, \texttt{vegetated mosaic}, \texttt{wetland} \\
\bottomrule
\end{tabularx}
\end{table}

\paragraph{Automated lexical audit ($\mathcal{A}$).}
Automated auditing computes $\mathcal{A}(\hat{L},\hat{E})=\mathbb{I}_{\mathrm{src}}(\hat{E})\cdot\mathbb{I}_{\mathrm{attr}}(\hat{E})\cdot\mathbb{I}_{\mathrm{cue}}(\hat{L},\hat{E})\in\{0,1\}$ using Table~\ref{tab:app-dict}. Term $\mathbb{I}_{\mathrm{src}}$ checks that $\hat{E}$ names optical/MSI and SAR sources. Term $\mathbb{I}_{\mathrm{attr}}$ splits $\hat{E}$ at sentence breaks (periods or semicolons) and \texttt{while}-clauses. It rejects clauses mixing optical-only tokens* solely with SAR sources, or SAR-only tokens* solely with optical sources. Term $\mathbb{I}_{\mathrm{cue}}$ checks that each predicted label $c\in\hat{L}$ matches an entry in $T_c$. Auditing checks attribution and cue coverage, while Section~\ref{app:human-audit} tests visual correctness through blinded experts.

\subsection{Double-Blind Human Audit}
\label{app:human-audit}

Two experts audited 100 validation patches stratified by class and outcome. Annotators inspected 400 randomized outputs, covering Joint SFT and EC-DPO (six views and blind), blinded to model identity, image access, labels, and accuracy. Because verbose formatting exposes the base model, they scored 100 zero-shot outputs separately (Table~\ref{tab:human-audit}). Viewing all six views, experts scored three rubrics:
\begin{enumerate}
\item \textbf{Scene Specificity (1--5 Likert scale):} spans five anchors: \textit{1 (generic)}, \textit{2 (vague)}, \textit{3 (moderate)}, \textit{4 (detailed)}, and \textit{5 (specific cross-view spatial and spectral cues)}. Generic text lacking spatial detail scores $1.26\text{--}1.27$ when blind.
\item \textbf{Physical Consistency (\texttt{plausible} vs.\ \texttt{implausible}):} checks cross-sensor attribution, flags modal swaps like NDVI under SAR, and verifies plausibility against all six views.
\item \textbf{Unsupported Cues (\texttt{none}, \texttt{one}, \texttt{multiple}):} flags phrases unsupported by the displayed images. Hallucination rate sums \texttt{one} and \texttt{multiple}.
\end{enumerate}
Inter-annotator agreement reaches $90.25\%$ ($\kappa=0.866$) on specificity, $94.00\%$ ($\kappa=0.873$) on consistency, and $94.25\%$ ($\kappa=0.895$) on hallucinations. EC-DPO raises mean specificity from 3.67 to \textbf{4.31} ($+0.64$, 95\% CI $[+0.50, +0.79]$). Physical consistency rises from $81.5\%$ to \textbf{95.5\%} ($+14.0\,\text{pp}$, 95\% CI $[+6.0, +22.0]\,\text{pp}$), while hallucinations drop from $18.5\%$ to \textbf{4.5\%} ($-14.0\,\text{pp}$, 95\% CI $[-22.0, -6.0]\,\text{pp}$). Without images, specificity falls to $1.26\text{--}1.27$ and hallucinations rise to $65.5\%\text{--}69.0\%$, confirming reliance on the available visual evidence.

\subsection{Primary Validation Results}
\label{app:confirmatory}

On validation, primary joint SFT reaches 0.8242 Micro F1 and 0.7561 Macro F1, alongside 0.4288 Exact Match and 0.9753 evidence audit. EC-DPO initializes from this checkpoint, training one epoch over 2,000 pairs ($\beta=0.02$, LR $5\times10^{-7}$, length-normalized log probabilities, gradient clip 1.0). Micro F1 rises to 0.8526, Macro F1 to 0.7783, Exact Match to 0.4720, and audit to 0.9829.

\subsection{Controlled Three-Seed LoRA Suite}
\label{app:seeds}

We run a three-seed suite (seeds 42, 123, 2026) under fixed batch order, separating visual tuning from language capacity. We compare language-only LoRA ($r=16$, 43.65M), rank-matched language LoRA ($r=19$, 51.84M), and joint LoRA ($r=16$, 51.35M). Moving 7.70M weights to visual blocks lifts Micro F1, from $0.8008\pm0.0037$ to $0.8309\pm0.0053$, and lifts \texttt{wetland} recall by $+10.94\,\text{pp}$ (Table~\ref{tab:per-seed-lora}, \mbox{Main Table~\ref{tab:training-ablation}a}).

\subsection{View Naming and Ordering}
\label{app:inputablation}

Main Table~\ref{tab:rendering-design}(D) tests view naming and ordering, keeping pixels fixed ($n=3{,}234$ patches). Replacing generic slots (\texttt{Image 1..6}, \texttt{D1}: 0.7884 Micro F1, 0.3541 Exact) with physical names (\texttt{A4/G1-d}) lifts Micro F1 to 0.7939, while Macro F1 hits 0.6995. Permuting order randomly under physical names (\texttt{D3}) drops Micro F1 to 0.7870, cutting Exact Match by $-0.0180$ (0.3578 to 0.3398), so fixed order anchors multi-view grounding.

\subsection{Leave-One-View-Out Ablation}
\label{app:lovo}

\begin{table*}[t]
\centering
\caption{Leave-one-view-out (LOVO) inference omission on the six-view joint SFT model ($3{,}234$ validation patches, 10,000 paired bootstrap resamples; $\Delta$ Micro computed prior to four-decimal rounding). Bold marks the highest value per column, and underlines mark the sharpest class-recall drops upon omitting True Color (\texttt{urban}) or SWIR (\texttt{wetland}).}
\label{tab:lovo}
\scriptsize
\setlength{\tabcolsep}{1.8pt}
\begin{tabular*}{\textwidth}{@{\extracolsep{\fill}}lrrrrrrrrrrr@{}}
\toprule
& \multicolumn{5}{c}{Overall Multilabel Metrics \& Prediction Density} & \multicolumn{6}{c}{Per-Class Recall Across Land-Cover Categories} \\
\cmidrule(lr){2-6} \cmidrule(lr){7-12}
Inference Condition & Micro F1 & $\Delta$ Micro (95\% CI) & Prec. & Recall & L/Patch & Urban & Forest & Cropland & Grass & Water & Wetland \\
\midrule
Intact (all 6 views) & 0.8242 & reference & 0.7743 & \textbf{0.8810} & 2.489 & 0.8282 & 0.9369 & 0.9070 & \textbf{0.7752} & 0.9334 & 0.7481 \\
Omit True Color & 0.7924 & $-0.0318$ $[-0.0372, -0.0265]$ & 0.7677 & 0.8188 & 2.333 & \underline{0.5583} & 0.8984 & 0.8576 & 0.6990 & 0.9142 & \textbf{0.8015} \\
Omit False Color & 0.8234 & $-0.0008$ $[-0.0037, +0.0022]$ & 0.7801 & 0.8719 & 2.445 & 0.8405 & 0.9316 & 0.9041 & 0.7492 & 0.9245 & 0.7099 \\
Omit SWIR & \textbf{0.8283} & $+0.0041$ $[+0.0016, +0.0068]$ & \textbf{0.7943} & 0.8653 & 2.383 & \textbf{0.8466} & 0.9268 & \textbf{0.9085} & 0.7283 & 0.9155 & \underline{0.6412} \\
Omit NDVI & 0.8210 & $-0.0031$ $[-0.0059, -0.0004]$ & 0.7812 & 0.8651 & 2.422 & 0.8016 & 0.9359 & 0.8793 & 0.7511 & 0.9373 & 0.7023 \\
Omit NDBI & 0.8233 & $-0.0009$ $[-0.0033, +0.0018]$ & 0.7838 & 0.8671 & 2.420 & 0.8180 & 0.9345 & 0.8942 & 0.7365 & \textbf{0.9398} & 0.7023 \\
Omit SAR & 0.8226 & $-0.0016$ $[-0.0040, +0.0007]$ & 0.7738 & 0.8779 & 2.481 & 0.8446 & \textbf{0.9379} & 0.9011 & 0.7644 & 0.9321 & 0.7328 \\
\bottomrule
\end{tabular*}
\end{table*}

Table~\ref{tab:lovo} expands Main Figure~\ref{fig:lovo-pr} with full per-class recall and 10,000-resample paired bootstrap CIs under leave-one-view-out omission. Because Red (B04) and NIR (B08) span multiple views, removing one derived composite barely shifts Micro F1. However, minority categories show strong reliance. Omitting True Color drops Blue (B02), cutting \texttt{urban} recall by $-26.99\,\text{pp}$. Omitting SWIR drops rare \texttt{wetland} recall by $-10.69\,\text{pp}$ ($74.81\%$ to $64.12\%$). Omitting NDVI ($-4.58\,\text{pp}$), False Color ($-3.82\,\text{pp}$), or NDBI ($-4.58\,\text{pp}$) also reduces \texttt{wetland} sensitivity.

\subsection{Visual Feature PCA and Cloud Saliency}
\label{app:feature-analysis}

\begin{figure*}[!b]
\centering
\includegraphics[width=0.90\textwidth]{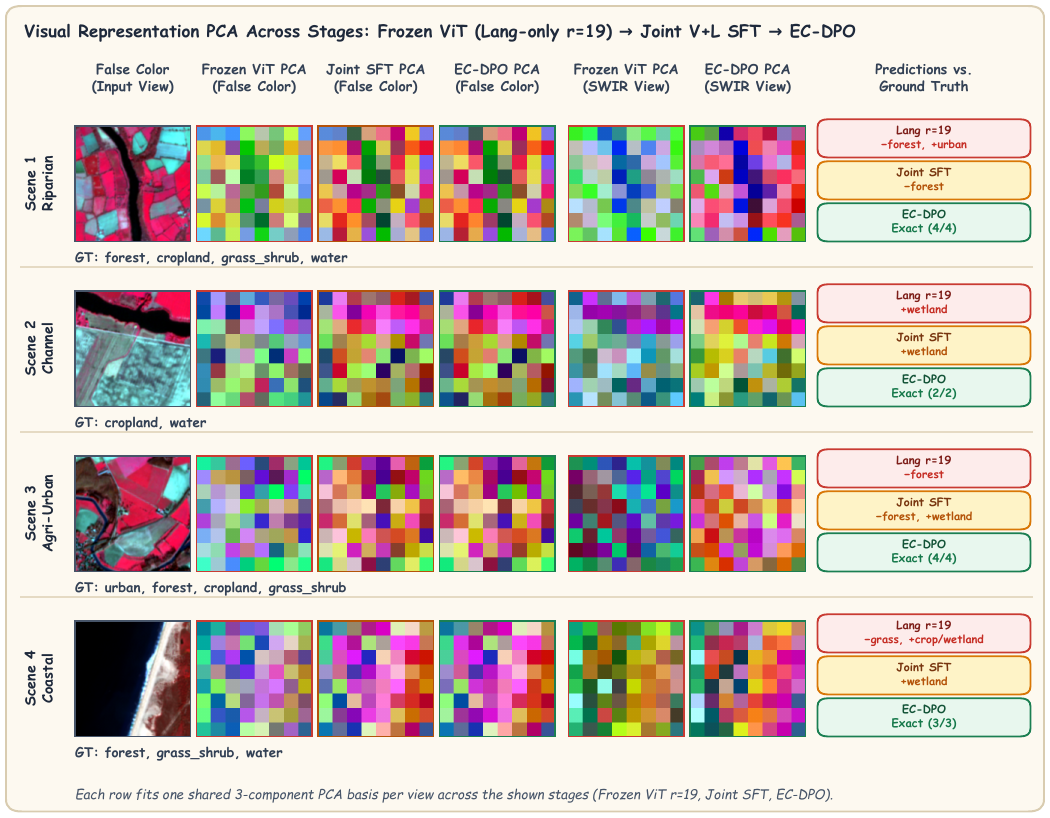}
\caption{Three-stage post-merger PCA comparison across four BigEarthNet-v2 validation scenes (Scenes~1--2 expand Main Figure~\ref{fig:qualitative-dpo}a). Each row fits one shared three-component SVD basis per view across the displayed stages (Frozen ViT $r=19$, Joint V+L SFT, and EC-DPO) with 2nd--98th percentile scaling. \texttt{FC}: False Color.}
\label{fig:app-pca-progression}
\end{figure*}

\begin{figure*}[t]
\centering
\includegraphics[width=0.96\textwidth]{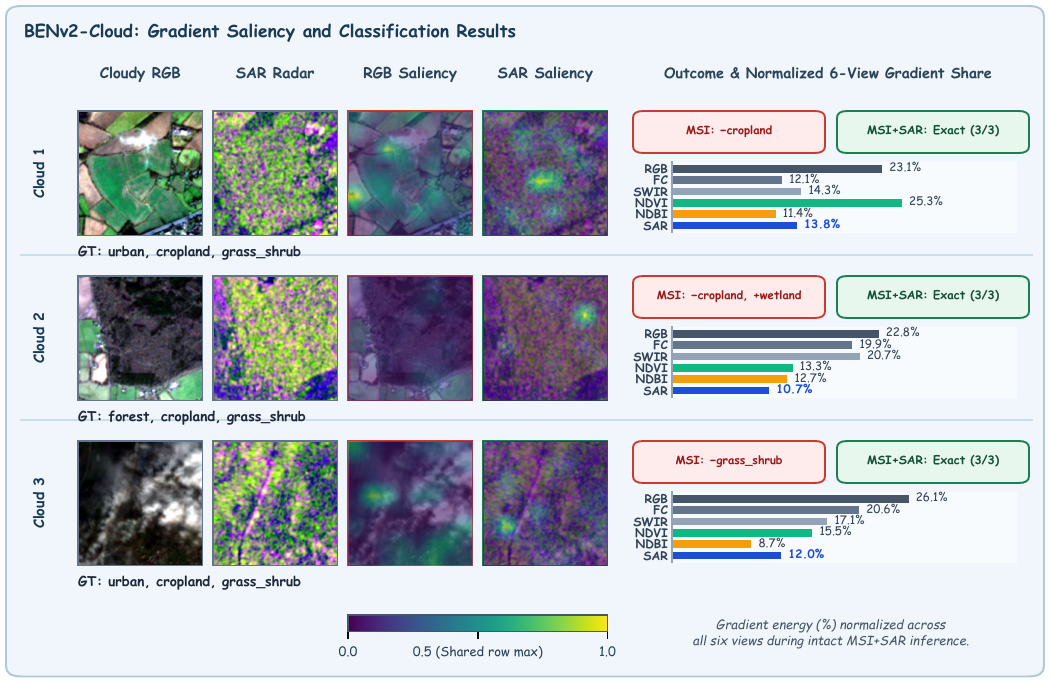}
\caption{Extended \texttt{BENv2-Cloud} gradient saliency maps across three occluded patches where MSI-only fails and intact MSI+SAR reaches Exact Match. RGB (\texttt{True color}) and SAR maps share one row maximum (\texttt{viridis} scale), and horizontal bars report the normalized six-view gradient energy distribution (\texttt{FC}: False Color).}
\label{fig:app-cloud-saliency}
\end{figure*}

Feature visualizations complement Main Figure~\ref{fig:qualitative-dpo}.
\begin{itemize}
\item \textbf{Multi-stage PCA evolution (Figure~\ref{fig:app-pca-progression}):} For four patches, we project $8\times8$ tokens per view onto one shared three-component SVD basis (Scenes~1--2 expand Main Figure~\ref{fig:qualitative-dpo}a). The panels compare False Color and SWIR projections across Frozen ViT ($r=19$), Joint SFT, and EC-DPO alongside their predicted class sets.
\item \textbf{Cloud-occluded SAR saliency (Figure~\ref{fig:app-cloud-saliency}):} Across three extra \texttt{BENv2-Cloud} scenes (beyond Main Figure~\ref{fig:qualitative-dpo}b), we sum $8\times8$ token gradient norms per view. Overlays share one row maximum (\texttt{viridis}); the bars report relative gradient magnitudes across the six views.
\end{itemize}

\subsection{Extended Qualitative Comparisons}
\label{app:qualitative}

\begin{figure*}[!p]
\centering
\includegraphics[width=0.94\textwidth]{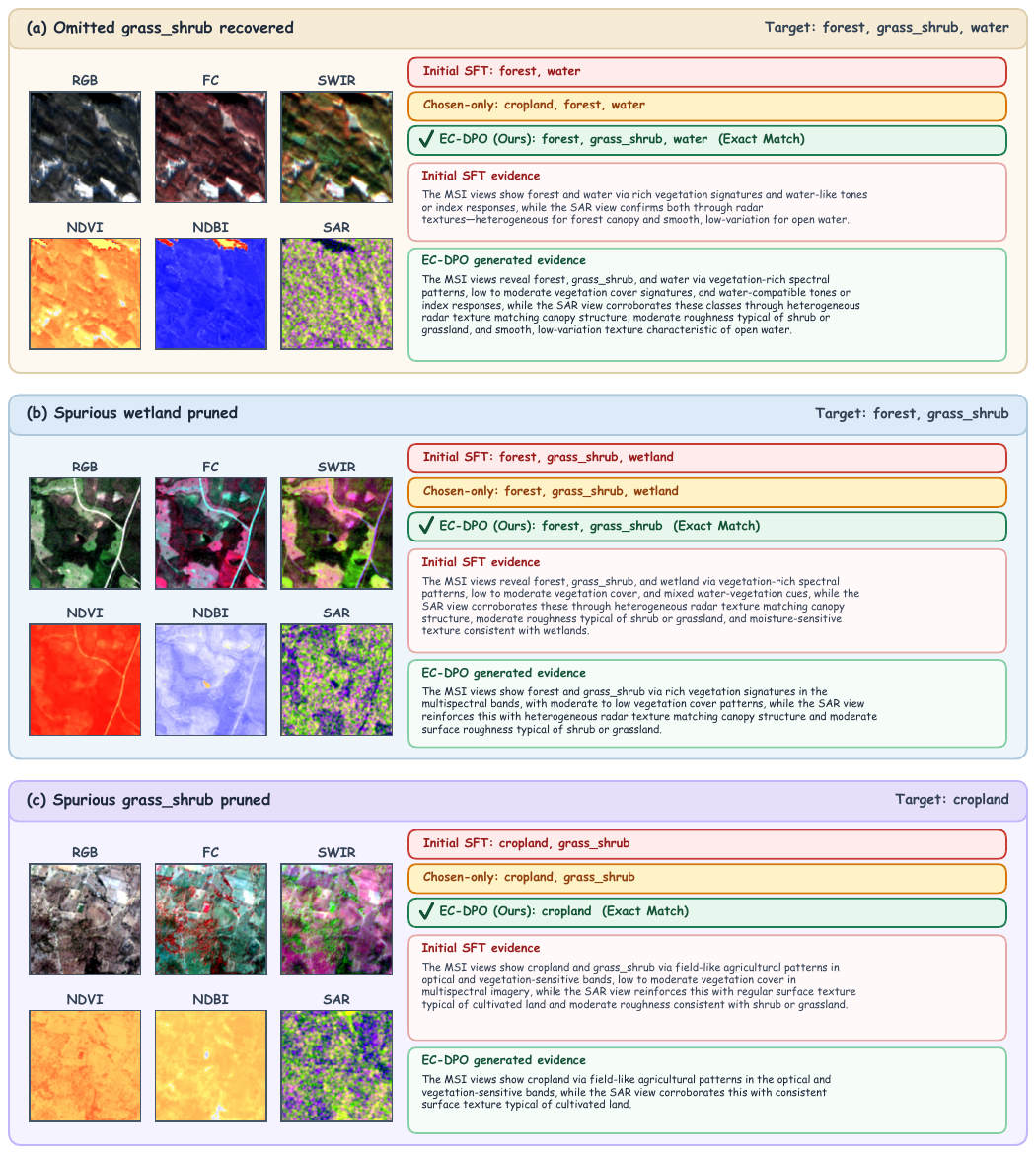}
\caption{Paired six-view qualitative comparisons contrasting Initial SFT, Chosen-only SFT, and EC-DPO predictions together with Initial SFT vs.\ EC-DPO sensor rationales across three validation scenes: (a)~recovery of omitted \texttt{grass\_shrub}, (b)~pruning of spurious \texttt{wetland}, and (c)~pruning of spurious \texttt{grass\_shrub}. \texttt{RGB}: True color; \texttt{FC}: False color.}
\label{fig:app-dpo-paired}
\end{figure*}

\begin{figure*}[!p]
\centering
\includegraphics[width=0.95\textwidth]{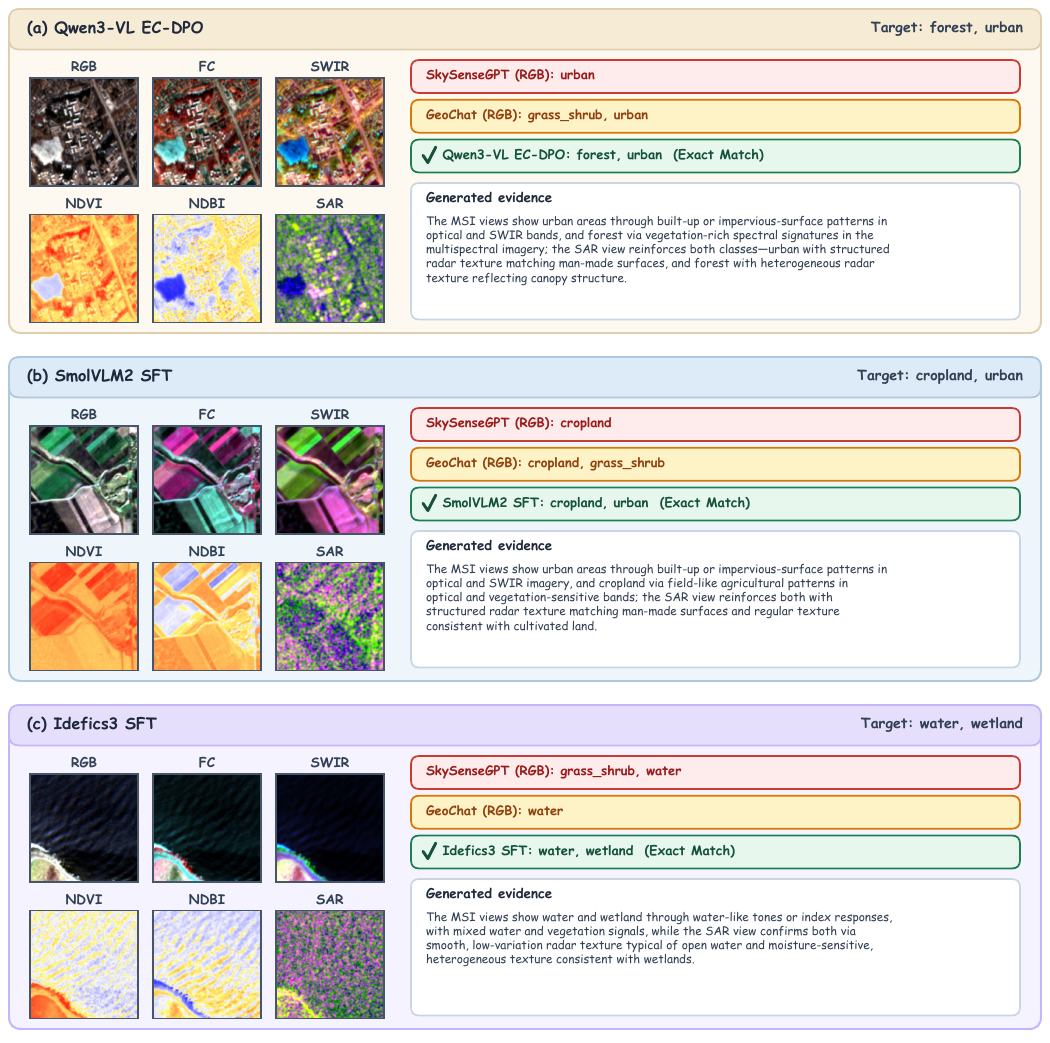}
\caption{Cross-backbone six-view qualitative comparison (Part~a): Qwen3-VL EC-DPO, SmolVLM2 SFT, and Idefics3 SFT achieve Exact Match across urban--forest, agricultural--urban, and coastal water--wetland scenes where RGB-only SkySenseGPT and GeoChat omit or hallucinate classes.}
\label{fig:app-cross-model-qualitative-a}
\end{figure*}

\begin{figure*}[!p]
\centering
\includegraphics[width=0.95\textwidth]{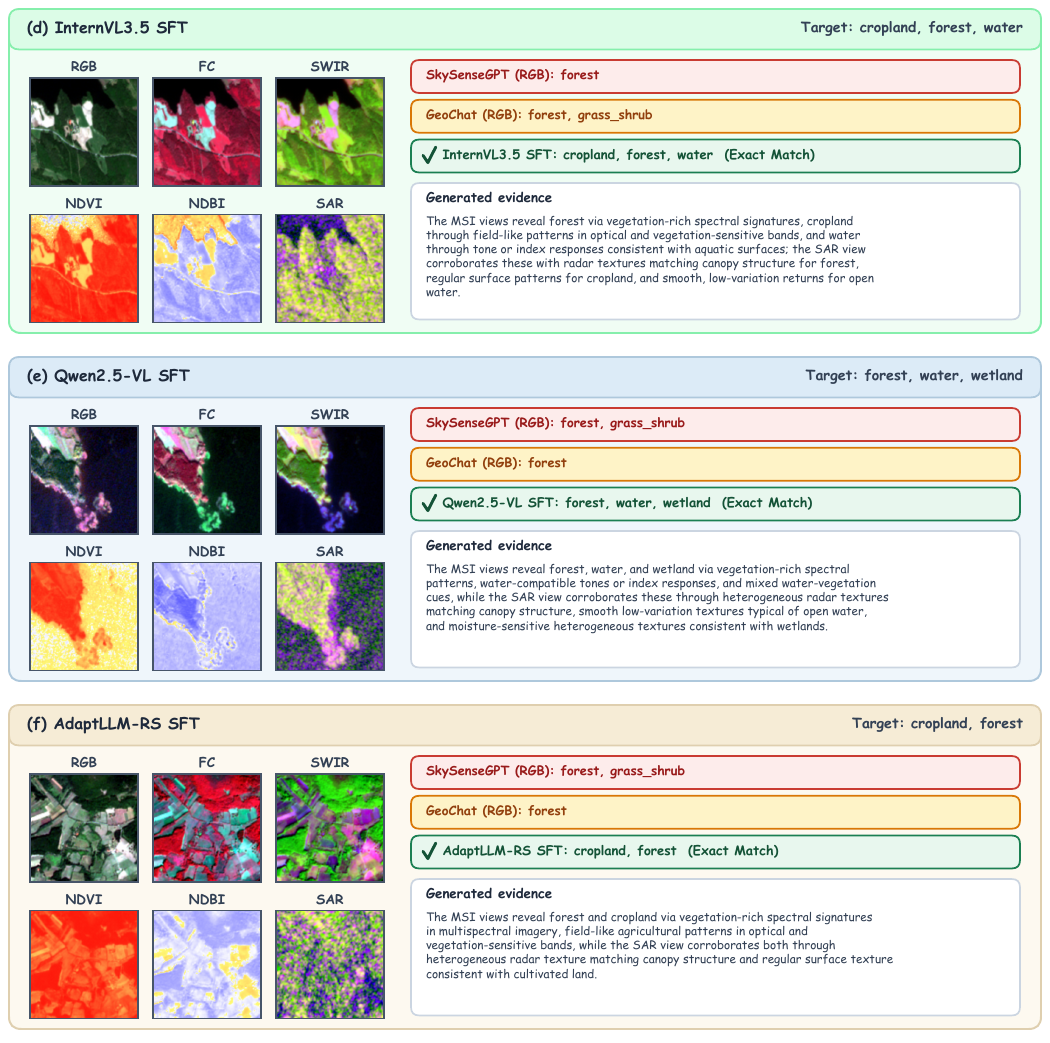}
\caption{Cross-backbone six-view qualitative comparison (Part~b): InternVL3.5 SFT, Qwen2.5-VL SFT, and AdaptLLM-RS SFT achieve Exact Match across cropland--forest--water, forest--water--wetland, and cropland--forest scenes alongside complete six-view sensor rationales.}
\label{fig:app-cross-model-qualitative-b}
\end{figure*}

\begin{figure*}[!p]
\centering
\includegraphics[width=0.95\textwidth]{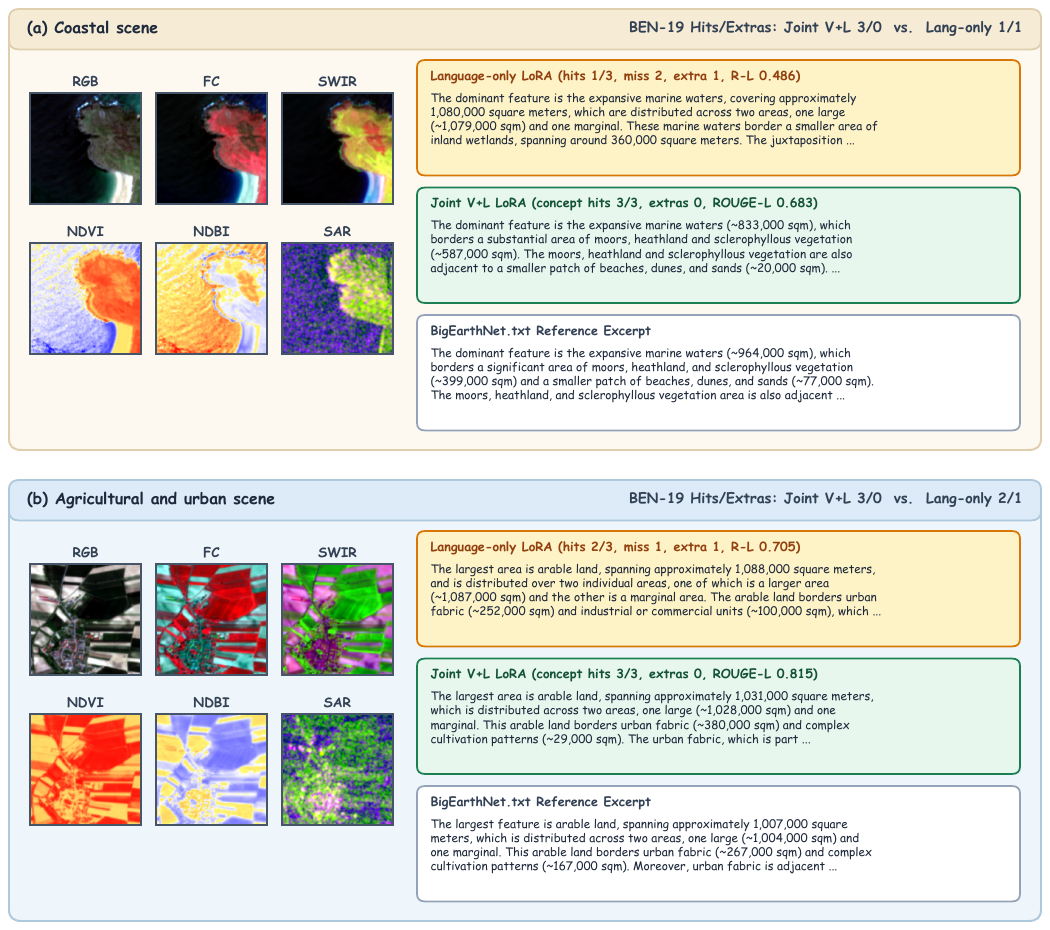}
\caption{BigEarthNet.txt dense captioning qualitative examples comparing Language-only LoRA, Joint V+L LoRA (Ours), and reference excerpts across coastal and agricultural/urban scenes with BigEarthNet-19 concept hits, extras, and ROUGE-L scores.}
\label{fig:app-caption-qualitative}
\end{figure*}

Figures~\ref{fig:app-dpo-paired}--\ref{fig:app-caption-qualitative} expand the qualitative comparisons across EC-DPO, VLM backbones, and dense captioning.
\begin{itemize}
\item \textbf{Paired EC-DPO examples (Figure~\ref{fig:app-dpo-paired}):} Six-view panels compare Initial SFT, Chosen-only SFT, and EC-DPO outputs, showing \texttt{grass\_shrub} recovery and spurious \texttt{wetland}/\texttt{grass\_shrub} pruning.
\item \textbf{Cross-backbone six-view transfer (Figures~\ref{fig:app-cross-model-qualitative-a}--\ref{fig:app-cross-model-qualitative-b}):} In these examples, six adapted VLMs reach Exact Match, while RGB-only SkySenseGPT and GeoChat miss or hallucinate classes.
\item \textbf{Dense captioning outputs (Figure~\ref{fig:app-caption-qualitative}):} Coastal and agro-urban BigEarthNet.txt scenes compare outputs, contrasting Language-only LoRA against Joint V+L LoRA and reference captions.
\end{itemize}

\subsection{Cross-Task Transfer: Flood Verification and Captioning}
\label{app:transfer}

\begin{table}[t]
\centering
\caption{Sen1Floods11 test evaluation ($178$ claims over $89$ patches, comprising $66$ water-positive vs.\ $112$ non-water claims with $1{:}1$ verification balance). Schema compliance is 1.0000 in all rows.}
\label{tab:app-sen1floods}
\footnotesize
\setlength{\tabcolsep}{1.2pt}
\begin{tabular*}{\columnwidth}{@{\extracolsep{\fill}}lrrrrr@{}}
\toprule
Setting & Flood F1 & Non-fld & Macro & Acc. & Verif. \\
\midrule
Qwen3-VL zero-shot & 0.5294 & 0.5699 & 0.5497 & 0.5506 & 0.5169 \\
\textbf{LoRA SFT, MSI+SAR} & \textbf{0.7714} & \underline{0.8519} & \textbf{0.8116} & \textbf{0.8202} & \textbf{0.8202} \\
LoRA SFT, MSI only & \underline{0.7072} & 0.6971 & 0.7022 & 0.7022 & 0.7022 \\
LoRA SFT, SAR only & 0.6604 & \textbf{0.8560} & \underline{0.7582} & \underline{0.7978} & \underline{0.7978} \\
LoRA SFT, no image & 0.0000 & 0.7724 & 0.3862 & 0.6292 & 0.6292 \\
\bottomrule
\end{tabular*}
\end{table}

\begin{table}[t]
\centering
\caption{BigEarthNet.txt dense captioning ($970$ test patches). We report BigEarthNet-19 concept F1, SacreBLEU, ROUGE-L without stemming, and mean token length.}
\label{tab:app-caption}
\footnotesize
\setlength{\tabcolsep}{1.5pt}
\begin{tabular*}{\columnwidth}{@{\extracolsep{\fill}}P{0.42\columnwidth}rrrr@{}}
\toprule
Adapter Setting & BEN-19 & BLEU & R-L & Tok. \\
\midrule
Base Qwen3-VL zero-shot & 0.0148 & 0.81 & 0.1177 & 179.8 \\
SFT without images (blind) & 0.2753 & 31.88 & 0.4421 & 104.0 \\
SFT Lang-only (6 views) & \underline{0.4270} & \underline{38.13} & \underline{0.5090} & 86.8 \\
\textbf{SFT Joint V+L (6 views)} & \textbf{0.5654} & \textbf{42.07} & \textbf{0.5337} & 90.6 \\
\bottomrule
\end{tabular*}
\end{table}

On Sen1Floods11 (Main Table~\ref{tab:transfer}a, Table~\ref{tab:app-sen1floods}), each patch renders four views: true color (\texttt{[B04, B03, B02]}), water SWIR (\texttt{[B08, B12, B03]}), NDWI ($(\text{B03}-\text{B08})/(\text{B03}+\text{B08})$), and SAR (\texttt{[VV, VH, VV-VH]}). Each scene pairs with two claims, \texttt{water} vs.\ \texttt{non-water}, under three output lines (\texttt{consistent}, \texttt{class}, \texttt{evidence}). This converts 337 train/val scenes into 674 claims, plus 178 claims from 89 test scenes, trained for 3 epochs (rank-16 LoRA, LR $10^{-4}$, $0.3\times$ visual scale). Water truth uses a $5\%$ pixel threshold (66 flood vs.\ 112 non-flood, $37.1\%$ prior), while claim consistency stays $1{:}1$ balanced (89 consistent, 89 inconsistent). Because SFT aligns its verification label with its prediction (\texttt{consistent} iff \texttt{class} matches the claim), class accuracy equals verification accuracy across all SFT rows ($0.8202$, $0.7022$, $0.7978$, $0.6292$). Zero-shot outputs instead self-contradict ($0.5506$ vs.\ $0.5169$). Blind no-image SFT predicts \texttt{non-water} everywhere, giving $\text{Flood F1}=0.0000$ and $112/178=62.92\%$ accuracy. Class and verification confusion counts ($\text{TP}/\text{FP}/\text{TN}/\text{FN}$) are $(45/59/53/21,\, 88/85/4/1)$ for zero-shot and $(54/20/92/12,\, 67/10/79/22)$ for MSI+SAR, while corresponding counts are $(64/51/61/2,\, 58/22/67/31)$ for MSI-only, $(35/5/107/31,\, 69/16/73/20)$ for SAR-only, and $(0/0/112/66,\, 56/33/56/33)$ for no-image.

On BigEarthNet.txt (Main Table~\ref{tab:transfer}b, Table~\ref{tab:app-caption}), Joint V+L LoRA trains for 1 epoch on 11,377 captions (LR $3\times10^{-5}$, $0.3\times$ visual scale), decoding greedily up to 256 tokens across 970 disjoint test scenes. Reference captions mix visible cover with unobservable metadata. Blind SFT reaches $31.88$ BLEU and $0.4421$ ROUGE-L, with concept F1 of $0.2753$ compared with $0.5654$ for Joint V+L. Figure~\ref{fig:app-caption-qualitative} shows example captions.

\begin{table*}[!t]
\centering
\footnotesize
\caption{Summary provenance for all baseline and adapted models evaluated on the 3,234-patch BigEarthNet-v2 validation split (\mbox{Main Table~\ref{tab:main}}).}
\label{tab:app-provenance}
\setlength{\tabcolsep}{2.5pt}
\begin{tabularx}{\textwidth}{@{}P{0.165\textwidth}P{0.345\textwidth}P{0.13\textwidth}X@{}}
\toprule
Tier / Model & Checkpoint Identifier & Input Interface & Adaptation \& Prompt / Decoding Spec \\
\midrule
Tier 1: TerraFM & \path{MBZUAI/TerraFM} & Native S1+S2 & Frozen backbone + 6-class linear head \\
Tier 1: CROMA & \path{antofuller/CROMA} & Native S1+S2 & Frozen backbone + 6-class linear head \\
Tier 1: ResNet50 & \path{torchvision.models.resnet50} & Six views & Frozen ImageNet backbone + 6-class head \\
Tier 1: Qwen3-VL probe & \path{Qwen/Qwen3-VL-8B-Instruct} & Six views & Mean-pooled frozen tokens + linear head \\
\midrule
Tier 2: SkySenseGPT & \path{ll-13/SkySenseGPT-7B-CLIP-ViT} & True color only & Zero-shot, \texttt{P-RGB}, greedy 160 tok. \\
Tier 2: GeoChat & \path{MBZUAI/geochat-7B} & True color only & Zero-shot, \texttt{P-RGB}, greedy 160 tok. \\
Tier 2: EarthDial-RGB & \path{akshaydudhane/EarthDial_4B_RGB} & Six views & Zero-shot, \texttt{View 1..6}+\texttt{P2} (\texttt{P-ED}), 160 tok. \\
Tier 2: EarthMind & \path{sy1998/EarthMind4B_multi} & Six views & Zero-shot, \texttt{<image>}+\texttt{P2}, 160 tok. (1 err.) \\
\midrule
Tier 3: Qwen3-VL-8B & \path{Qwen/Qwen3-VL-8B-Instruct} & Six views & Zero-shot / 1-ep.\ Lang LoRA (\texttt{P2}, 96 tok.) \\
Tier 3: Qwen2.5-VL-7B & \path{Qwen/Qwen2.5-VL-7B-Instruct} & Six views & Zero-shot / 1-ep.\ Lang LoRA (\texttt{P2}, 96 tok.) \\
Tier 3: AdaptLLM-RS-3B & \path{AdaptLLM/remote-sensing-Qwen2.5-VL-3B-Instruct} & Six views & Zero-shot / 1-ep.\ Lang LoRA (\texttt{P2}, 96 tok.) \\
Tier 3: InternVL3.5-8B & \path{OpenGVLab/InternVL3_5-8B-HF} & Six views & Zero-shot / 1-ep.\ Lang LoRA (\texttt{P2}, 96 tok.) \\
Tier 3: Idefics3-8B & \path{HuggingFaceM4/Idefics3-8B-Llama3} & Six views & Zero-shot / 1-ep.\ Lang LoRA (\texttt{P2}, 96 tok.) \\
Tier 3: SmolVLM2-2.2B & \path{HuggingFaceTB/SmolVLM2-2.2B-Instruct} & Six views & Zero-shot / 1-ep.\ Lang LoRA (\texttt{P2}, 96 tok.) \\
\midrule
Tier 4: Qwen2.5-VL-7B & \path{Qwen/Qwen2.5-VL-7B-Instruct} & Six views (grid) & 2-ep.\ Joint V+L LoRA, \texttt{P2}, greedy 96 tok. \\
Tier 4: Qwen3-VL-8B & \path{Qwen/Qwen3-VL-8B-Instruct} & Six views (grid) & 2-ep.\ Joint V+L LoRA (+EC-DPO), \texttt{P2}, 96 tok. \\
\bottomrule
\end{tabularx}
\end{table*}

\subsection{Counterfactual Sensor Interventions}
\label{app:image-dependence}

Main Table~\ref{tab:image-dependence} evaluates counterfactual interventions across two validation cohorts. The first cohort samples 800 clear-sky patches (seed 42). The second selects 1,000 cloudy or shadowed patches (\texttt{BENv2-Cloud}) via the \texttt{contains\_cloud\_or\_shadow} quality flag. Blind text-prior runs drop all image parts, keeping labels and prompts fixed. For donor swaps, each target pairs 1-to-1 with a donor patch, via linear sum assignment over 1,000 trials. This matching minimizes label Jaccard overlap without self-pairs. Rows (a)--(e) and (g) use primary Joint SFT, scoring 0.8180 Micro F1 in clear sky. Row~f uses a seeded Joint V+L SFT reference (0.8162 Micro F1, 0.4088 Exact Match). There, True Color and SAR stay intact while False Color, SWIR, NDVI, and NDBI come from the donor.

As an additional clear-sky control, replacing all six views with a donor drops Micro F1 to 0.4806 and changes $93.63\%$ of label sets. In clear sky, swapping SAR alone alters $5.75\%$ of predictions. Swapping the four derived spectral views lowers Micro F1 from 0.8162 to 0.7287 and Exact Match to 0.2213. Under that four-view swap, \texttt{wetland} recall drops by $39.40\,\text{pp}$ ($87.88\%$ to $48.48\%$), and \texttt{water} recall drops by $36.08\,\text{pp}$ ($93.81\%$ to $57.73\%$). On \texttt{BENv2-Cloud}, cloud occlusion raises SAR donor sensitivity to $22.80\%$ and SAR removal sensitivity to $23.00\%$. Meanwhile, SAR-only inference retains 0.6463 \texttt{water} F1 and 0.6328 \texttt{forest} F1.

\subsection{EC-DPO Preference Controls}
\label{app:dpo}

Main Table~\ref{tab:training-ablation}(b) reports matched 2,000-pair controls initialized from Joint V+L SFT (0.8242 Micro F1, 0.4288 Exact Match). Each pair draws a multi-label patch ($|L^\star|\ge 2$), dropping one positive label ($L_- = L^\star\setminus\{c\}$) to form a hard negative while prompts and preferred targets stay fixed. EC-DPO preserves full evidence on the rejected side ($E_-=E^\star$), decoupling classes from cues, whereas the coupled control strips the dropped class's clause ($E_-=E^\star\setminus T_c^\star$). Over Joint V+L SFT, EC-DPO lifts Micro F1 by $+0.0284$, to $0.8526$, and boosts Exact Match by $+0.0432$ to $0.4720$ ($p<0.001$). It also lifts \texttt{wetland} recall by $+5.29\,\text{pp}$ to $0.8010$. Against chosen-only SFT on identical targets, EC-DPO gains $+0.0189$ Micro F1, $+0.0608$ Exact Match, and $+6.05\,\text{pp}$ \texttt{wetland} recall.

\subsection{Baseline Provenance and Prompt Interfaces}
\label{app:baseline-provenance}

Table~\ref{tab:app-provenance} summarizes models, inputs, and decoding setups across Tiers~1--4 of Main Table~\ref{tab:main} ($n=3{,}234$ patches). On one patch, \texttt{EarthMind-4B} hits an empty-image-feature error, so we keep it in the denominator as invalid. Tier~3 runs six backbones under one shared chat wrapper, reaching $0.7681$ Micro F1 on 1-epoch Qwen3-VL-8B. With the native grid processor, the language-only reference reaches $0.7939$ Micro F1 ($+0.0258$; Main Table~\ref{tab:rendering-design}, \texttt{G1-d}).

\paragraph{Single-View and Native Multi-View Prompts (\texttt{P-RGB}, \texttt{P-ED}).}
True-color models (SkySenseGPT, GeoChat) see only true color under prompt~\texttt{P-RGB}, which enforces the six-class candidate list and two-line contract. Models must cite visible RGB cues and omit unrendered SWIR, NDVI, NDBI, or SAR features. EarthDial-RGB prepends indexed tags (\texttt{View~1: <image>} to \texttt{View~6: <image>}) before prompt~\texttt{P2} (\texttt{P-ED}).

\paragraph{Canonical Six-View Prompt (\texttt{P2}).}
Tier~3 and Tier~4 multisensor models use this prompt and two-line schema:
\begin{list}{}{\setlength{\leftmargin}{0.75em}\setlength{\rightmargin}{0.2em}\setlength{\topsep}{3pt}\setlength{\partopsep}{0pt}}
\item\footnotesize\raggedright\ttfamily
Question: What land-cover types are visible in this remote-sensing patch?\\
Candidate classes: urban, forest, cropland, grass\_shrub, water, wetland.\\
Use the named optical, derived-index, and radar views as complementary evidence sources.\\
Image 1: optical true-color rendered~view.\\
Image 2: optical vegetation false-color rendered~view.\\
Image 3: optical SWIR rendered~view.\\
Image 4: optical NDVI derived-index rendered~view.\\
Image 5: optical NDBI derived-index rendered~view.\\
Image 6: SAR radar rendered~view.\\
Answer with exactly two lines:\\
class: <comma-separated labels from the~candidates>\\
evidence: <short sensor-grounded reason>
\end{list}

\end{document}